\documentclass[acmlarge]{acmart}
\makeatletter
\newcommand{\confshort}{\acmConference@shortname}
\newcommand{\conffull}{\acmConference@name}
\newcommand{\confdate}{\acmConference@date}
\newcommand{\confloc}{\acmConference@venue}
\AtBeginDocument{
  \fancypagestyle{firstpagestyle}{
    \fancyhead{}%
    \fancyfoot[C]{}%
  }
  \fancyhf{}
  \fancyhead[LO]{\@headfootfont\shorttitle}%
  \fancyhead[RE]{\@headfootfont\@shortauthors}%
  \fancyhead[LE]{\@headfootfont\footnotesize \confshort, \confdate, \confloc}%
  \fancyhead[RO]{\@headfootfont\footnotesize \confshort, \confdate, \confloc}%
  \fancyfoot[C]{}%
}
\makeatother
\acmBooktitle{\conffull\@ (\confshort), \confdate, \confloc}

\AtBeginDocument{%
  }

\copyrightyear{2026}
\acmYear{2026}
\setcopyright{cc}
\setcctype{by}
\acmConference[FAccT '26]{The 2026 ACM Conference on Fairness, Accountability, and Transparency}{June 25--28, 2026}{Montreal, QC, Canada}
\acmBooktitle{The 2026 ACM Conference on Fairness, Accountability, and Transparency (FAccT '26), June 25--28, 2026, Montreal, QC, Canada}
\acmDOI{10.1145/3805689.3812290}
\acmISBN{979-8-4007-2596-8/2026/06}

\usepackage{graphicx}
\usepackage{booktabs}
\usepackage{amsmath}
\usepackage{mathtools}
\usepackage{amsthm}
\usepackage{color}
\usepackage[colorinlistoftodos]{todonotes}

\theoremstyle{definition}
\newtheorem{definition}{Definition}

\newtheoremstyle{examplestyle}  % name
  {}{}                          % space above/below
  {}                            % body font (empty = normal)
  {}                            % indent
  {\itshape}                    % head font — italic
  {.}                           % punctuation after head
  { }                           % space after head
  {}                            % head spec

\theoremstyle{examplestyle}
\newtheorem*{example}{Example}

\newcommand{\Lagr}{\mathcal{L}}
\usepackage{amsmath,amsfonts,bm}
\usepackage{multirow}
\usepackage{makecell}
\usepackage{subcaption}
\usepackage[font={small}]{caption}
\usepackage{wrapfig}
\def\eqref#1{equation~\ref{#1}}
\usepackage{amsmath}
\usepackage{bbm}

\newcommand{\calD}{\mathcal{D}}

\newcommand{\calR}{\mathcal{R}}

\newcommand{\calH}{\mathcal{H}}

\newcommand{\bx}{\mathbf{x}}

\newcommand{\bw}{\mathbf{w}}

\newcommand{\bq}{\mathbf{q}}
\newcommand{\bs}{\mathbf{s}}

\newcommand{\E}{\mathbb{E}}

\usepackage{amsthm}

\DeclareMathOperator*{\argmax}{argmax}

\usepackage{tikz}
\usetikzlibrary{arrows}

\usepackage{booktabs}


\begin{document}
\title{Rashomon Sets and Model Multiplicity in Federated Learning}

\author{Xenia Heilmann}
\affiliation{%
  \institution{Institute of Computer Science, Johannes Gutenberg University}
  \city{Mainz}
  \country{Germany}}
\email{xenia.heilmann@uni-mainz.de}
\author{Luca Corbucci}

\affiliation{%
  \institution{Fondazione Bruno Kessler}
  \city{Trento}
  %\state{Ohio}
  \country{Italy}
}
\email{lcorbucci@fbk.eu}
\author{Mattia Cerrato}
\affiliation{%
  \institution{Institute of Computer Science, Johannes Gutenberg University}
  \city{Mainz}
  \country{Germany}}
\email{mcerrato@uni-mainz.de}

\begin{abstract}
The Rashomon set captures the collection of models that achieve near-identical empirical performance yet may differ substantially in their decision boundaries. Understanding the differences among these models, i.e., their multiplicity, is recognized as a crucial step toward model transparency, fairness, and robustness, as it reveals decision boundary instabilities that standard metrics obscure. 
However, the existing definitions of Rashomon set and multiplicity metrics assume centralized learning and do not extend naturally to decentralized, multi-party settings like Federated Learning (FL). In FL, multiple clients collaboratively train models under a central server’s coordination without sharing raw data, which preserves privacy but introduces challenges due to heterogeneous client data distribution and communication constraints. 
In this setting, choosing a single ``best'' model may homogenize predictive behavior across diverse clients, amplify biases, or undermine fairness guarantees.
In this work, we provide the first formalization of Rashomon sets in FL.
First, we adapt the Rashomon set definition to FL, distinguishing among three perspectives: (I)~a \textbf{global} Rashomon set defined over aggregated statistics across all clients, (II)~a $t$-agreement Rashomon set for which a ratio of $t$ clients need to satisfy the same constraints, and (III)~\textbf{individual} Rashomon sets specific to each client’s local distribution.
Second, we show how standard multiplicity metrics can be estimated under FL’s privacy constraints. 
Finally, we introduce a multiplicity‑aware FL pipeline and conduct an empirical study on standard FL benchmark datasets. Our results demonstrate that all three proposed federated Rashomon set definitions offer valuable insights, enabling clients to deploy models that better align with their local data, fairness considerations, and practical requirements.
\end{abstract}
\maketitle

\section{Introduction}
\begin{figure}
\includegraphics[width=0.9\linewidth]{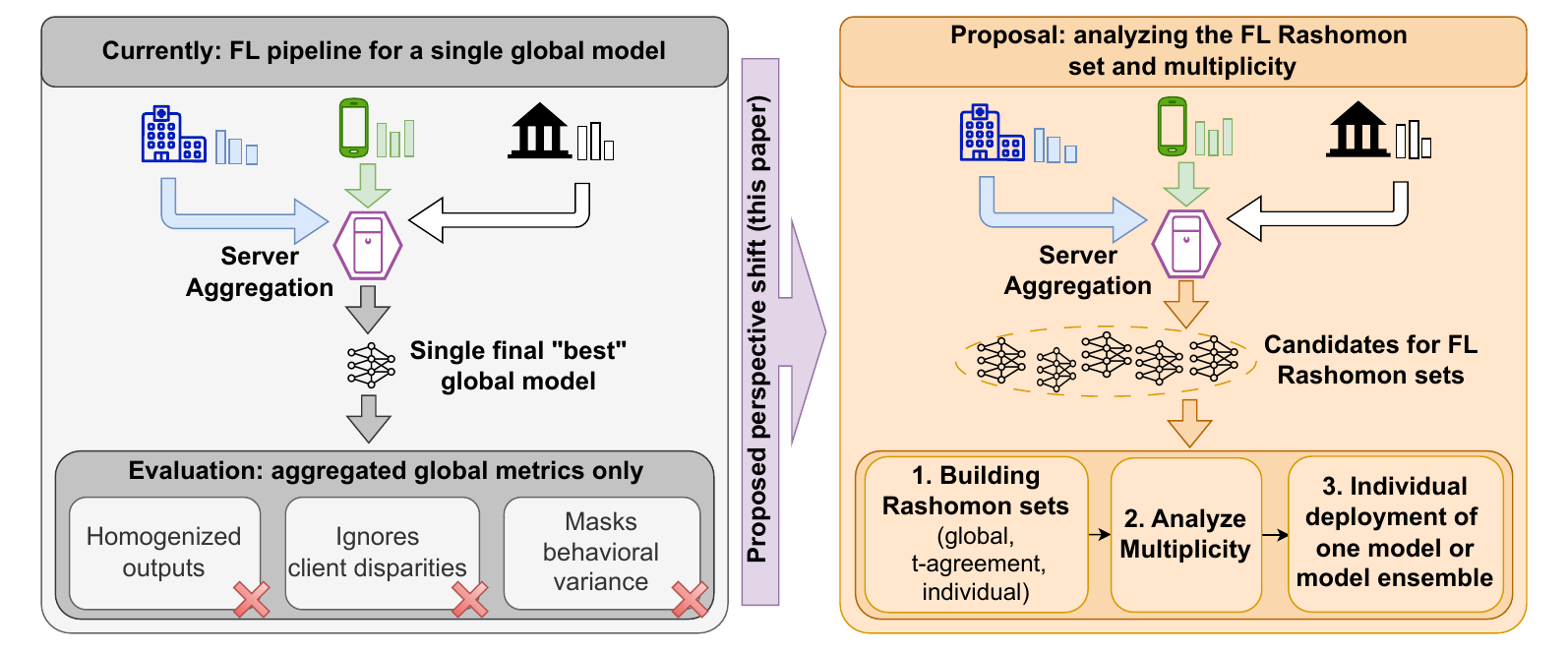} 
\caption{A paradigm shift needed in FL: the old ``single-best'' model hides significant differences in behavior across clients, obscuring perspectives that vary for different subsets of data. Exploring the Rashomon set enables a more accountable and transparent understanding of model performance.}
\label{fig:main_fig}
\end{figure}
Designing a Machine Learning (ML) algorithm requires making numerous decisions. Most of these decisions need to be taken well before model training actually starts, let alone deployment.
Leaving aside deliberation over the data collection procedure, decisions need to be made about preprocessing techniques, model family, hyperparameters (e.g., in neural networks, learning rate, batch size, local epochs), random seeds, model evaluation, early stopping criteria, and more. Each of these choices can be addressed in various plausible ways; each configuration of these choices yields distinct models that may perform similarly on the training data. These similar, yet distinct models are often described as the Rashomon set of models and the phenomenon as the Rashomon effect~\cite{breiman2001statistical}. 

The models in the Rashomon set may appear indistinguishable when evaluated using performance metrics such as test accuracy. However, they can hide different perspectives learned from the underlying data distributions. Selecting a model over another in the Rashomon set, solely based on similar performance metrics, constitutes an arbitrary decision that can have significant downstream implications. For instance, the chosen model may exhibit poorer fairness properties~\cite{fairness_ml} than the others in the Rashomon set for a specific group of people in the dataset, or it may yield explanations that are less robust or less interpretable~\cite{xai_robustness}. The research field of model multiplicity~\cite{black2022multiplicity} here seeks to understand and quantify this divergence within Rashomon sets, highlighting how different design choices during the model training can lead to conflicting predictions for the individual data samples~\cite{ganesh2025systemizingmultiplicitycuriouscase}.

While these concerns are relevant in centralized settings, where a central server holds the entire dataset, they have not yet been explored and studied in Federated Learning (FL)~\cite{mcmahan2023communicationefficient}.
FL is a decentralized, multi-party ML setting in which a central server orchestrates, for multiple rounds, the training of a model across decentralized clients (e.g., a smartphone, a hospital, a bank, etc.). Once selected, the clients train a model on their own private data and send the resulting model to the server, which aggregates all received models into a single global one. The core idea of FL is that training data remain on the clients’ devices, preserving privacy. At the same time, clients benefit from the federation's collective knowledge to train a more generalizable model. The outcome of this multi-round process is a single, final model that is then distributed to all clients. 
This multi-party setting makes many of the design choices discussed above even more opaque, and the space of potential arbitrary decisions grows with each participating client. For example, data collection practices can differ across clients, feature selection may be performed locally through heterogeneous pipelines, hyperparameter tuning is often handled in a decentralized manner~\cite{khodak2021federated}, and client participation varies over time due to availability constraints. When taking part in the FL training, clients invest computation time, and thus expect a good global model as a reward. However, this is not necessarily the case; previous research~\cite{corbucci2025benefits, chang2023biaspropagationfederatedlearning} has shown how FL can lead to a strong homogenization of the predictive output across all clients, which may impact clients with ``non-typical'' data more than others. In particular, the standard FL evaluations based on averaged global metrics can hide substantial variation in model behavior ~\cite{corbucci2025benefits} and predictive outcomes across different clients.

In this context, a principled analysis of model multiplicity could be beneficial. Previous research \cite{Rudin2024AmazingTC} has argued that the existence of Rashomon sets is not necessarily a source of arbitrariness but rather a space of possibilities, where, for instance, it may be possible to obtain interpretable (i.e., simple, simulatable \cite{lipton2017mythosmodelinterpretability}) or fair models at little cost of model accuracy. 
The issue is, once more, perspective: it is far from straightforward to understand how to compute Rashomon sets and model multiplicity metrics in a federated, distributed setting.
Computing the Rashomon set globally on aggregated statistics could risk missing individual or minority perspectives on, for instance, which models should belong in the Rashomon set, or which data should be considered when establishing the predictive performance.
To the best of our knowledge, there is no established procedure or algorithm that tackles the computation of either Rashomon sets or predictive multiplicity metrics in FL. 
Our main aim with the present contribution is to tackle this problem while keeping in mind the challenge of perspective: in practice, we develop definitions and algorithms that are sensitive to the problem of data heterogeneity across clients. 

More broadly, as illustrated in Figure~\ref{fig:main_fig}, our findings point to the need for a paradigm shift in FL. We find that relying on a single ``best'' model is insufficient to ensure accountability, robustness, and fairness in settings where heterogeneous clients with diverse objectives and data distributions contribute to training.
Therefore, we deem a principled exploration of defining the Rashomon set as essential to uncover differences in model behavior and understand how algorithmic variance impacts individual clients or groups of clients. While characterizing a Rashomon set incurs non-trivial computational costs, these costs must be weighed against the higher risk of deploying an arbitrary single ``best'' model that may perform well on aggregate yet fail silently for minority clients or underrepresented groups.

In this work, we lay the foundation for analyzing Rashomon sets in FL. Our contributions are summarized as follows:
\begin{itemize}
    \item We formalize the definition of the Rashomon set in a decentralized setting, introducing \textbf{Global}, \textbf{Individual}, and \textbf{$t$-agreement Rashomon sets}.
    \item We propose and adaption of existing predictive multiplicity metrics into FL settings, while respecting its constraints.
    \item We propose an FL pipeline and an open-source implementation\footnote{\href{https://github.com/xheilmann/FederatedLearningMultiplicity}{https://github.com/xheilmann/FederatedLearningMultiplicity}}, conduct an empirical study to illustrate the differences among the three Rashomon set definitions in practice, and demonstrate a potential application. 
\end{itemize}
In Table~\ref{tab:symbols} in Appendix~\ref{app:symbols}, we report a list of the symbols used in the paper to increase clarity and readability.

\section{Background}\label{sec:background}
\subsection{Federated Learning}\label{sec:FL}

FL~\cite{mcmahan2023communicationefficient} is an approach to training ML models that keeps the data required for the learning process decentralized. In FL, a set of clients $\mathcal{C}$ collaboratively trains a global model $h$ for a certain number of rounds $R$. The process is orchestrated by a server $S$, which in each round $r \in \{1,\dots,R\}$ selects a subset $\psi \subseteq  \mathcal{C}$ of clients to run the training and then aggregates their model updates. Once selected, each client $c$ receives a model from the server and executes $\Gamma$ local training epochs on its dataset. The server then aggregates the models received from the clients. We exemplify here two algorithms for this: FedSGD~\cite{mcmahan2023communicationefficient} assumes that each client $c \in \psi$ executes a single gradient descent update before sharing the gradients $\hat{g}_c = \nabla \Lagr(\hat{h}_c, b_i)$ of the local model $\hat{h}_c$ on the batch $b_i$. The server aggregates the updates $h^{r+1} = h^r - \eta \sum_{c\in \psi} \frac{n_c}{n} \hat{g}_c$ where $n_c$ is the amount of samples of client $c$ and $n$ is the total number of samples.
The second algorithm we briefly discuss is FedAVG~\cite{mcmahan2023communicationefficient}, a more advanced algorithm that allows each client to perform the model update $\hat{h}_{c}^{(e+1)} = \hat{h}_{c}^{(e)} - \eta \nabla \Lagr(\hat{h}_c^{(e)}, b_i); e\in\{0,\dots,\Gamma-1\}$
for $\Gamma$ epochs before sharing the model $\hat{h}_c:=\hat{h}_c^{(\Gamma)}$ with the server that will aggregate all the received local models $\hat{h}_{c}$ producing the model $h^{r+1}$ for the next round $h^{r+1} \leftarrow \sum_{c\in \psi} \frac{n_c}{n} \hat{h}_{c}$.

After training has finished, FL model evaluation can be conducted in several ways. For centralized evaluation, the server holds a globally representative test set and evaluates the global model directly at the end of training. For distributed evaluation, the evaluation process mirrors the structure of an FL training round, but evaluation clients only return metric summaries, i.e., loss, accuracy, or counts, to the server. The server then combines these per-client statistics by applying an evaluation aggregation mechanism $f_E$, typically a weighted average, to obtain a global estimate of the metric.

Based on the number of clients involved in the training and on their availability, we can distinguish between cross-silo and cross-device FL~\cite{huang2022crosssilofederatedlearningchallenges}. When we have ten to hundreds of always available clients, we are in a cross-silo scenario. On the other hand, when the number of clients grows to thousands, and they are only available under specific conditions, we are in a cross-device scenario. 

Despite the goal of protecting clients' privacy, FL is vulnerable to multiple privacy attacks~\cite{Zhao_2025, boenisch2023curious} and is usually used alongside Differential Privacy (DP)~\cite{10.1007/11787006_1} to formally guarantee a bound on the clients' privacy risk (see Appendix~\ref{app:diffpriv_explanation} for a definition of Differential Privacy).

\subsection{The Rashomon Effect in Machine Learning}
\label{sec:rashomon}

The term Rashomon effect originates from Akira Kurosawa's 1950 film ``Rashomon'' and describes the phenomenon in which several observers generate mutually inconsistent, yet internally coherent, descriptions of the same event. In ML, the term was adopted to describe the existence of sets of models with similar error rates~\cite{breiman2001statistical}. These sets of models are now referred to by various names, including Rashomon sets, sets of good models, $\epsilon$-Rashomon sets, or sets of competing models. In this paper, we adopt the term \textit{Rashomon set}. We define a Rashomon set following the broad framework of Ganesh et al. ~\cite{ganesh2025systemizingmultiplicitycuriouscase}. Let $\Delta$ be a set of metric functions, where each $\delta_i \in \Delta$ takes as input two models and quantifies their difference according to a specific metric. Each metric is associated with a corresponding threshold $\mathcal{E}$. If for every $\delta_i\in \Delta$ the measured difference between two models falls within the corresponding threshold $\epsilon_i \in \mathcal{E}$, the models are considered similar and belong to the same Rashomon set. 
Formally, we have:
\begin{definition}[Rashomon Set~\cite{ganesh2025systemizingmultiplicitycuriouscase}]
Two models $h_1, h_2$ belong to the same Rashomon set under performance constraints $(\Delta, \mathcal{E})$ if they exhibit similar performance for every metric in the given performance constraints, i.e.: 
\begin{align}\label{eq:RashomonSet}
    \delta_i (h_1, h_2)\leq \epsilon_i \quad \forall (\delta_i, \epsilon_i) \in (\Delta, \mathcal{E})
\end{align}
\end{definition}
\noindent In essence, the Rashomon set contains all models whose expected risk for each metric lies within $\epsilon$ of the optimal value. When this set is large, many qualitatively different models satisfy the same performance criterion. 

Identifying all models in the Rashomon set is computationally infeasible. To address this, the full Rashomon set $\mathcal{R}$ is typically approximated by a subset of $m$ models referred to as \textit{empirical Rashomon set}, defined as 
$$\mathcal{R}^m(\mathcal{H},  (\Delta, \mathcal{E}), h_{\bw^\ast}) \;\triangleq\;
\bigl\{\, h_1,\dots,h_m \in \mathcal{H} ;
h_i \in \mathcal{R}(\mathcal{H}, (\Delta, \mathcal{E}), h_{\bw^\ast}) \,\,\forall i \in [m] \,\bigr\}$$
where $h_{\bw^\ast}$ is the baseline model, $\mathcal{H}$ the set of all models in the hypothesis space and $m$ the size of the empirical Rashomon set~\cite{hsu2024rashomongb}. The most common approach to approximate this set is via a re-training strategy, where models are trained from scratch with different random initializations~\cite{KulynychDPMultiplicity, semenova2019study}. Then, models deviating from the constraints in Equation~\ref{eq:RashomonSet} are rejected, and the process repeats until $m$ models are found.
Alternatives to re-training include RashomonGB for gradient boosting algorithms~\cite{hsu2024rashomongb} and dropout-based approaches, where neurons are randomly removed during training~\cite{hsu2024dropout}. 
Once a Rashomon set is available, practitioners can select a model or ensemble based on secondary criteria such as interpretability~\cite{dwivedi2023explainable}, fairness~\cite{fairness_ml}, or robustness~\cite{freiesleben2023beyond}. This approach enables informed model selection, allowing practitioners to choose models that meet specific requirements while understanding the broader implications of their choices~\cite{he2023visualizing, EerlingsAiSpectra}.

\subsection{Multiplicity}
Multiplicity~\cite{breiman2001statistical} refers to situations where models exhibit similar overall performance but diverge in behavior. As these divergences can take many forms, multiplicity is studied in several subfields such as predictive multiplicity~\cite{pmlr-v119-marx20a}, explanation multiplicity~\cite{gunasekaran2024explanation, Watson_2022_WACV}, fairness multiplicity~\cite{wang2021directional, islam2021can, coston2021characterizing}, and allocation multiplicity~\cite{jain2025allocation}, among others. Focusing on the first three, predictive multiplicity occurs when models within the Rashomon set produce conflicting predictions. Explanation multiplicity arises when multiple explanations for the same sample appear equally plausible. Fairness multiplicity occurs when models satisfy different fairness criteria or meet the same criteria to varying degrees. A formal definition, consistent with the Rashomon set definition, is as follows:
\begin{definition}[Model Multiplicity~\cite{ganesh2025systemizingmultiplicitycuriouscase}]
Two models $h_1, h_2$ exhibit multiplicity under performance constraints $(\Delta, \mathcal{E})$ and multiplicity constraint $(\delta^M, \epsilon^M)$, if they achieve similar performance for every metric in the given performance constraints, but differ according to the multiplicity metric, i.e.: 
\begin{align}\label{eq:multiplicity}
    \delta_i (h_1, h_2)\leq \epsilon_i \quad \forall (\delta_i, \epsilon_i) \in (\Delta, \mathcal{E}) \quad \text{and} \quad \delta^M(h_1,h_2)>\epsilon^M.
\end{align}
\end{definition}

\section{Related Work}\label{sec:related}

To the best of our knowledge, this is the first work tackling the definition of both Rashomon sets and multiplicity metrics in FL. There has, however, been substantial work on tailoring FL models to heterogeneous client requirements or local data distributions through personalization strategies \cite{tan2022towards}. These range from post-hoc adaptation of a trained global model \cite{li2021ditto, li2020fedprox, wang2020optimizing} to methods that directly learn personalized models \cite{smith2017federated, lin2020ensemble, diao2020heterofl}. Some personalization research also targets specific dimensions, such as fairness. For instance, Fair Hypernetworks support FL settings where different clients optimize for different fairness metrics \cite{fairPFLdifferentfairnessmetrics}.
Our work on multiplicity in FL is complementary to, rather than a replacement for, these personalization efforts. Personalization addresses an optimization problem: how to produce a model that performs well under each client's local data distribution. Multiplicity addresses a diagnostic question: among the models that achieve comparable performance, how much do their predictions disagree, and for whom? These are orthogonal, if complementary, concerns.
We believe the diagnostic perspective is particularly relevant in FL because the conditions that drive multiplicity (heterogeneous data distributions, variation in feature relevance across clients, and underspecification from limited local data) are inherent to federated settings. Multiplicity metrics can therefore surface properties of the learning problem itself: high predictive multiplicity for certain individuals or subgroups may signal that the federation's data does not sufficiently constrain the solution space for those cases, a finding that no single model, personalized or otherwise, would reveal on its own.
Multiplicity is also significant for contestability and accountability in high-stakes FL deployments. When multiple equally valid models disagree on a prediction for a specific individual, this disagreement is material: it means the outcome depends on an arbitrary model selection choice rather than on the evidence alone \cite{Rudin2024AmazingTC}. Surfacing this information allows clients and affected individuals to identify when outcomes are robust and when they are contestable, a consideration that applies regardless of whether the deployed model has been personalized.
In this work, we focus on multiplicity in non-personalized FL. However, understanding how personalization strategies affect multiplicity, both at the global model level and across individual clients, is an interesting direction for future research.

\section{Characterizing the Rashomon Set in Federated Learning}
\label{sec:characterizing}
Adapting the definition and the evaluation of the Rashomon set to a decentralized setting remains an open challenge. Any meaningful definition should account for the client-server architecture inherent to FL, capturing both a global federation-level perspective and a local, client-level perspective. In the following definitions, we will analyze the hypothesis space $\calH$ over federated models $h_i$ after federated training has finished, so after round $R$. Based on this, an off-the-shelf approach would be to compute the Rashomon set over these candidate FL models independently on each client and take their intersection. However, computing the client-level Rashomon sets does not automatically generalize to an overall globally valid Rashomon set over aggregated metrics. On the other hand, defining a Rashomon set based on aggregated server-level statistics can hide client-specific heterogeneity, potentially overlooking fairness and individualized performance considerations. To address these limitations, we introduce three definitions for Rashomon sets in FL settings. Each definition is valid for the complete federation, highlights a distinct aspect of the FL setting, supports different evaluation criteria, and enables unique opportunities for later model selection. In short, we propose (1) a global definition, (2) a $t$-agreement definition, and (3) an individual, client‑wise definition.
We define these three propositions in the following in a theoretical manner, illustrate them with a running example, and provide an empirical study in Section~\ref{sec:empiricalstudy}.

\subsection{Global Rashomon Set}
During FL model training, two evaluation methodologies are commonly employed. In the first, a subset of clients from the client pool evaluates the model on their entire local data (cross-device setting). In the second, each client holds separate validation and test subsets, distinct from the local training data, for evaluating the model (cross-silo setting).
Incorporating these evaluation techniques into the Rashomon‑set definition requires that the inequality in Equation~\ref{eq:RashomonSet} holds for the subset of clients $c$ belonging to the evaluation client set $\mathcal{C}_E$. Concretely, each client $c \in \mathcal{C}_E$ computes the metric differences $\delta_{i} (h_1, h_2)_c$ individually. These client-level measurements $\delta_i(h_1,h_2)_1,\dots,\delta_i(h_1,h_2)_{|\mathcal{C}_E|} $ are then aggregated using the underlying FL evaluation aggregation algorithm $f_E$, yielding a single federation-level measurement: $f_E(\delta_i(h_1,h_2)_1,\dots,\delta_i(h_1,h_2)_{|\mathcal{C}_E|} )$. 
This leads to the following definition:
\begin{definition}[Global Rashomon set]\label{def:global-rashomon}
Two models $h_1, h_2$ belong to the same \textbf{global Rashomon set} under performance constraints $(\Delta, \mathcal{E})$ if they exhibit similar performance for \textbf{every aggregated} metric in the given performance constraints, i.e.: 
\begin{align}\label{eq:globalRashomonSet}
   f_E(\delta_i(h_1,h_2)_1,\dots,\delta_i(h_1,h_2)_{|\mathcal{C}_E|} )\leq \epsilon_i \quad \forall (\delta_{i}, \epsilon_i) \in (\Delta, \mathcal{E}).
\end{align}
\end{definition}

\noindent A key advantage of this definition is its alignment with the standard FL evaluation practices. However, models that satisfy the definition for the aggregated statistics across the evaluation clients may still fail to meet them on individual, randomly selected clients. This first definition demonstrates how standard FL aggregation masks model instability, serving as a contrast to the more granular definitions that incorporate client-specific perspectives that follow in the next section.

\begin{example}\label{ex:running}
To build intuition on how the three definitions relate to Rashomon set construction, consider a federation with evaluation clients $\mathcal{C}_E$ and $|\mathcal{C}_E| = 3$ and models $h_1, h_2, h_3 \in \calH$ which were trained by the federation for $R$ rounds. Suppose the performance constraint is defined by a single metric $\delta$ (e.g., accuracy difference) with the constraint $\epsilon = 0.002$, and that the per-client measurements are:
\begin{center}
\begin{tabular}{lccc}
\toprule
 client& $c=1$ &  $c=2$ &  $c=3$  \\
\midrule
$n_c$ & 1000 & 500 & 500 \\
$\delta(h_1, h_2)_c$ & 0.001 & 0.001 & 0.003\\
$\delta(h_2, h_3)_c$ & 0.006 & 0.002 & 0.004\\
\bottomrule
\end{tabular}
\end{center}
Suppose the evaluation aggregation function $f_E$ is the weighted average. Then, this yields 
 \begin{align*}&f_E(\delta(h_1,h_2)_1,\delta(h_1,h_2)_2, \delta(h_1,h_2)_3 ) =\frac{1000}{2000}*0.001 +  \frac{500}{2000}*0.001 + \frac{500}{2000}*0.003  = 0.0015\leq \epsilon,\text{ and} \\&f_E(\delta(h_2,h_3)_1,\delta(h_2,h_3)_2,\delta(h_2,h_3)_3) =\frac{1000}{2000}*0.006 +  \frac{500}{2000}*0.002 + \frac{500}{2000}*0.004  = 0.0045\geq \epsilon.
 \end{align*}
 Thus, $h_1$ and $h_2$ are in the same global Rashomon set while $h_2$ and $h_3$ are not.

\end{example}

\subsection{$\mathbf{t}$-Agreement Rashomon Set}
To address the limitation of Definition~\ref{def:global-rashomon} in capturing individual client perspectives, we introduce a threshold $t \in (0,1]$ together with an agreement strategy. This $t$ determines the ratio of clients for which the individual measurements $\delta_{i} (h_1, h_2)_c$ must satisfy the performance constraints to be considered part of the same Rashomon set. By varying $t$, the influence of outliers that aggregated statistics may hide can be mitigated.

\begin{definition}[$t$-agreement Rashomon set]\label{def:threshold-rashomon}
Two models $h_1, h_2$ belong to the same \textbf{$t$-agreement Rashomon set} under performance constraints $(\Delta, \mathcal{E})$ if they exhibit similar performance for every metric in the given performance constraints \textbf{for at least a ratio $\mathbf{t}$ of clients in the evaluation set}, i.e.: 
\begin{align}\label{eq:thresholdRashomonSet}
\frac{ \Big |\{c\in\mathcal{C}_E|\delta_{i} (h_1, h_2)_c\leq \epsilon_i\,\,\, \forall (\delta_{i}, \epsilon_i) \in (\Delta, \mathcal{E}) \} \Big|}{|\mathcal{C}_E|} \geq t
\end{align}
\end{definition}

\noindent Applying a small value for $t$ results in Rashomon sets where the performance constraints are only fulfilled by some evaluating clients. On the contrary, choosing $t > 0.5$ ensures that the Rashomon set contains only models that locally satisfy the performance constraints across the majority of evaluating clients. When $t=1.0$, this definition reduces to the intersection of all local Rashomon sets. 
We note that for a high value of $t$, the resulting $t$-agreement Rashomon set may be empty. This proves a fundamental incompatibility among client data distributions, suggesting that no single model can simultaneously satisfy the specified performance constraints across evaluating clients.
While this definition takes into account individual clients in the evaluation client set, it remains limited to the subset of clients $\mathcal{C}_E$. This motivates a further, complete client‑centric definition.

\begin{example}[
]
Building onto the earlier example, notice that $\delta (h_1, h_2)_c \leq 0.002$ is fulfilled for $c\in\{1,2\}$ whereas  $\delta (h_2, h_3)_c \leq 0.002$ is  only fulfilled for $c=2$. Hence, $h_1,h_2$ belong to the same $t$-agreement Rashomon set if $t \leq \frac{2}{3}$, whereas $h_2,h_3$ only belong to the same $t$-agreement Rashomon set if $t \leq \frac{1}{3}$. Requiring majority agreement ($t > 0.5$) would thus reflect that $h_1,h_2$ are in the same $t$-agreement Rashomon set, while $h_2,h_3$ are not. 

\end{example}

\subsection{Individual Rashomon Set}
Adapting the Rashomon set definition to a client‑wise perspective assumes that each client $c$ in an FL system maintains its own local Rashomon set. In this setting, no single global Rashomon set exists. Formally, each client $c \in \mathcal{C}$ holds a set of FL models that satisfies the performance constraints in Inequality~\ref{eq:RashomonSet}. However, two variations can be considered. First, the performance constraint $(\Delta, \mathcal{E})$ can be defined globally across all clients, thereby coupling them through a common FL objective (shared constraints). Second, each client can specify its own constraints $(\Delta_c, \mathcal{E}_c)$, yielding fully individualized Rashomon sets (individual constraints).
The first option reflects the server-based nature of FL, where a central server can coordinate evaluation metrics across clients, while the second respects complete client autonomy. 

The following definition adopts the former option, i.e., a shared global constraint with client‑specific evaluation, which can be easily extended to individual constraints by replacing $(\Delta, \mathcal{E})$ with the individual version $(\Delta_c, \mathcal{E}_c)$.

\begin{definition}[Individual Rashomon set]\label{def:client-rashomon}
For each client $c\in\mathcal{C}$, two models $h_1, h_2$ belong to the same \textbf{individual Rashomon set} under \textbf{globally defined} performance constraints $(\Delta, \mathcal{E})$ if they exhibit similar performance on client $c$`s evaluation data for every metric in the given performance constraints, i.e.: 
\begin{align}\label{eq:individualRashomonSet}
 \delta_{i} (h_1, h_2)_c\leq \epsilon_i \quad \forall (\delta_{i}, \epsilon_i) \in (\Delta, \mathcal{E}) 
\end{align}
\end{definition}
\noindent This definition balances central oversight with client-level customization. Adopting the global constraints prevents clients from deviating too much from the overall FL objective, while each client maintains its own Rashomon set that reflects its data distribution.
\begin{example}
    We again build on the earlier example. Here, each client evaluates the models independently, and thus Rashomon sets can vary across clients. Thus, for client $1$ $h_1,h_2$ are in the same individual Rashomon set; for client $2$, $h_1, h_2$ and $h_2, h_3$ are in the same individual Rashomon sets, and for client $3$, none are in the same sets. 
\end{example}

\section{Predictive Multiplicity Metrics} \label{sec:multimetrics}
Having established a theoretical foundation for defining Rashomon sets in FL, we now develop predictive multiplicity metrics for Rashomon set evaluation in FL. Recent work has introduced a variety of such metrics~\cite{hsu2022_rashomoncapacity,pmlr-v119-marx20a, watson2023predictive, long2023arbitrariness}, each capturing a different dimension of model variability within the Rashomon sets. However, it is non-trivial to apply these metrics in FL. Their canonical formulation does not respect FL privacy constraints and assumes centralized access to data. Thus, the challenge is to determine how these metrics can be adapted to FL while respecting privacy. In the following, we propose strategies for adapting existing metrics to the FL setting. For each strategy, we highlight the difficulties stemming from the adaptation.

\subsection{Score-based Metrics}
Predictive multiplicity metrics are generally grouped into score-based and decision-based metrics.
Score-based metrics, such as Rashomon Capacity (RC)~\cite{hsu2022_rashomoncapacity}, Viable Prediction Range (VPR)~\cite{watson2023predictive}, and standard deviation~\cite{long2023arbitrariness}, focus on the spread of model output scores (we report definitions in Appendix~\ref{app:metrics}). Their computation typically relies on estimating quantiles or the cumulative distribution function of these metrics. Extending this to FL is not straightforward. The simplest approach would require clients to share the metric for each datapoint with the server, enabling a centralized estimation of quantiles and cumulative distribution. Such communication exposes client-level distributions and does not preserve the privacy of the clients. Thus, it can only be applied in settings where clients trust the central server. 
\subsubsection{Privacy-Preserving Modifications}\label{sec:privpres}
We sketch here different privacy-preserving techniques that can relax this assumption. One option is to employ differentially private histograms to communicate the cumulative distribution of scores~\cite{kellaris2013practical, dwork2006calibrating,hay2009boosting}. The procedure works as follows. i) the server pre-defines a set of buckets; ii) each client bins the calculated metric values for its local data points accordingly and randomly perturbs the number of points in each bucket, so as to preserve differential privacy; iii) the noisy client-level histograms are sent to the server; and iv) the server aggregates them to obtain a global approximation of the cumulative score distribution. Global quantiles can then be estimated directly from this aggregated distribution. We revisit this technique in Section~\ref{sec:empiricalstudy} and Appendix~\ref{app:diffpriv}. A key challenge is that some histogram bins may be empty, and the addition of noise using standard DP noise mechanisms can lead to a poor approximation of the underlying distribution. Solutions for distributed communication protocols in a two-server model for computing sparse histograms have been proposed~\cite{bell2022distributedsparsehist, braun2024malicious}. However, these approaches rely on two non-colluding servers, which does not apply to general FL settings. Also, secure aggregation methods could be applied on top of histograms, such as the SecAgg protocol~\cite{bonawitz2016practical}, which is based on Shamir’s secret sharing. %Other options include adding evaluation techniques derived from verification scemes~\cite{}. 

\subsection{Decision-based Metrics} 
We define decisions here as either a threshold score or a score vector after $\argmax$. For binary classification, a model produces a score $s$, the decision is obtained by $\mathbbm{1}[s > \tau]$, where $\tau$ is a threshold and $\mathbbm{1}[\cdot]$ is the indicator function.
For a $d$-class classification problem with $d > 2$, the model returns a score vector $\bs \in \Delta_d$, and the corresponding  decision is $\argmax\limits_{i \in [d]} [\bs]_i$. In this section, we use $\calD$ to denote the evaluation dataset, $h_{\bw^*}$ to denote a pre-trained baseline model, and $h$ is a model in the Rashomon set $\calR$. Compared to score-based metrics, which require approximation of the metrics in FL, most decision-based metrics can be computed exactly. We now describe three such metrics: ambiguity, discrepancy, and disagreement.

\subsubsection{Ambiguity} 
Ambiguity~\cite{pmlr-v119-marx20a} measures the proportion of examples for which some models $h$ in the Rashomon set $\calR$ assign a different label from the one of the baseline $h_{\bw^*}$\footnote{An implementation for ambiguity and discrepancy can be found at \url{https://github.com/charliemarx/pmtools}.\label{footnote1}}. Formally:
\begin{align}\label{eq:ambiguity}
\alpha(\calD) &\triangleq \frac{1}{|\calD|} \sum\limits_{\bx_i\in\calD} \max\limits_{h\in \calR} \mathbbm{1} \left[ \argmax h(\bx_i) \neq \argmax h_{\bw^*}(\bx_i) \right].
\end{align}
Here, a high ambiguity value indicates that many examples get assigned conflicting predictions across all models, while small values indicate fewer conflicting predictions. 

In FL, however, $\calD$ is partitioned across $C$ clients. Concretely, we have $\calD_1, \dots,\calD_C$ datasets and their union would result in $\calD$. Each client $c\in\mathcal{C}$ can therefore compute only a local variant of $\alpha(\calD)$, namely 
 \begin{align}
     \alpha(\calD_c) &\triangleq \frac{1}{|\calD_c|} \sum\limits_{\bx_i\in\calD_c} \max\limits_{h \in \calR} \mathbbm{1} \left[ \argmax h(\bx_i) \neq \argmax h_{\bw^*}(\bx_i) \right]
 \end{align}
To recover the global ambiguity, clients communicate $\alpha(\calD_c)$ and the number of datapoints $|\calD_c|$ to the server. Then, the server can calculate a weighted aggregation similar to the one computed in the FedAVG algorithm~\cite{mcmahan2023communicationefficient}:
 \begin{align*}
     \alpha(\calD) &= \sum\limits_{c \in \mathcal{C}}\frac{|\calD_c|}{|\calD|} \alpha(\calD_c) =\sum\limits_{c \in \mathcal{C}}\frac{|\calD_c|}{|\calD|} \frac{1}{|\calD_c|}  \sum\limits_{\bx_i\in\calD_c} \max\limits_{h \in \calR} \mathbbm{1} \left[ \argmax h(\bx_i) \neq \argmax h_{\bw^*}(\bx_i) \right]
 \end{align*}
which is equal to Equation~\ref{eq:ambiguity}. 
To avoid revealing client-specific dataset sizes, the server may fix an evaluation set size for all the clients. This ensures that each client reports metrics computed over an identical number of samples. Alternatively, instead of applying a weighted average, the scores $\alpha(\calD_c)$ can be aggregated using the same secure aggregation method applied during the FL pipeline to aggregate evaluation metrics.

\subsubsection{Discrepancy} 
Discrepancy~\cite{pmlr-v119-marx20a} captures the maximum proportion of predictions that could change when replacing the baseline model with any other in the Rashomon set\textsuperscript{\ref{footnote1}}. Using the same notion as above, the mathematical definition follows as:
\begin{align}
\gamma(\calD) &\triangleq \max\limits_{h \in \calR} \frac{1}{|\calD|} \sum\limits_{\bx_i\in\calD}  \mathbbm{1} \left[ \argmax h(\bx_i) \neq \argmax h_{\bw^*}(\bx_i) \right]
\end{align}
Here, a low discrepancy value means that model predictions are similar to the baseline, while a high discrepancy value indicates the opposite. 
We adapt this metric to the FL setting by calculating $\gamma(\calD_c)$ locally on each client $c \in \mathcal{C}$ and then taking the weighted maximum over all $\gamma(\calD_c)$. This gives us the maximum number of predictions that could change:
\begin{align*}
    \gamma(\calD) &= \max\limits_{c \in \mathcal{C}} \frac{|\calD_c|}{|\calD|}\gamma(\calD_c)
    = \max\limits_{c \in \mathcal{C}}\frac{|\calD_c|}{|\calD|}\max\limits_{h \in \calR} \frac{1}{|\calD_c|} \sum\limits_{\bx_i\in\calD_c}  \mathbbm{1} \left[ \argmax h(\bx_i) \neq \argmax h_{\bw^*}(\bx_i) \right]
\end{align*}
Again, we can fix an evaluation set size to mask the number of datapoints on each client.

\subsubsection{Disagreement}
Disagreement~\cite{KulynychDPMultiplicity} is defined using the notion of probability and quantifies the extent to which models within a Rashomon set yield divergent predictions on a given instance. Formally, it is defined as
\begin{align}
    \mu(\bx_i)\triangleq2\text{Pr}\{\mathbbm{1}\left[ h_1(\bx_i) > \tau\right] \neq \mathbbm{1} \left[h_2(\bx_i)>\tau\right]; h_1, h_2 \in \mathcal{R}\}.
\end{align}
where $h_1, h_2$ denote any two models in the Rashomon set. This metric can be calculated in closed form over binary classifiers~\cite{KulynychDPMultiplicity}. Since this metric is defined for every datapoint $\bx_i$, its evaluation in FL is similar to one of the score-based metrics. Specifically, all clients compute $\mu(\bx_i)$ for each datapoint in their local data $\mathcal{D}_c$, then, in a naive implementation, they share all results with the server. Since this direct sharing does not preserve the privacy of client data, the methods described in Section~\ref{sec:privpres} can be used on top of the local distributions before sharing the results with the server. 

\begin{figure}
    \includegraphics[width=0.7\linewidth]{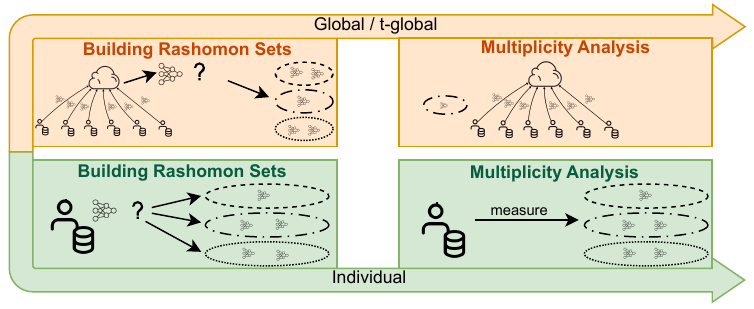}
    \caption{Complete FL pipeline for integration of Rashomon sets and multiplicity analysis}
    \label{fig:pipeline}
\end{figure}

\section{Multiplicity-aware FL Pipeline}\label{sec:pipeline}

Having established a formal framework for defining Rashomon sets in FL and adapting multiplicity metrics accordingly, we now introduce an end-to-end pipeline for integrating Rashomon set analysis into FL. The pipeline is visualized in Figure~\ref{fig:pipeline} and consists of three stages: \textbf{\textit{(i)}} Rashomon Set Candidate Generation, \textbf{\textit{(ii)}} Rashomon Set Construction, and \textbf{\textit{(iii)}} Multiplicity Analysis. Our pipeline requires at most two extra communication rounds beyond standard FL training.

\subsection{Rashomon Set Candidate Generation} In the first stage, the server orchestrates the training of multiple FL models to serve as candidates for the Rashomon sets. In general, this step mirrors standard FL pipelines (see Section~\ref{sec:FL}) with the only distinction that it produces a collection of federated models rather than a single one. To achieve this, the most straightforward approach is to adapt the re-training strategy to the FL context, for example, by varying the client sampling strategy, the number of rounds, or the fraction of participating clients for each FL training process. Although the re-training introduces significant computational overhead, we adopt it as our baseline because it remains the standard and theoretically sound approach for Rashomon set exploration in the centralized literature~\cite{ganesh2025systemizingmultiplicitycuriouscase}. To mitigate the computational overhead of re-training, which is exacerbated in the FL setting, we use simple neural network architectures. This choice is motivated not only by efficiency but also by recent findings in the multiplicity literature~\cite{Rudin2024AmazingTC}, which indicate that simpler models often reside within the Rashomon set and can achieve performance comparable to more complex architectures.

\subsection{Rashomon Set Construction} In the second stage, candidate models are evaluated and filtered to construct the Rashomon set. For the global and $t$-agreement Rashomon sets, all clients in the evaluation set evaluate all candidate models on their local data and share the results with the server, which aggregates them using the predefined aggregation function from the previous step. Then, the server determines membership in each Rashomon set relative to a baseline model and communicates the resulting assignments back to the clients. For the individual Rashomon sets, evaluation and filtering occur locally, and the results are not shared with the server. Note that this stage can also be merged with the first one when the full set of clients remains available throughout training and when their local training, validation, and test splits remain stable over time. At the end of this stage, every client holds a valid Rashomon based on either the global, $t$-agreement, or individual definition.

\subsection{Multiplicity Analysis} In the final stage, the multiplicity metrics introduced in Section~\ref{sec:multimetrics} are computed on the constructed Rashomon sets. In the global and $t$-agreement settings, clients compute the metrics locally and transmit them to the server, which aggregates the results as described in Section~\ref{sec:multimetrics} and returns them to the clients. These aggregated metrics can then support downstream tasks such as ensemble construction or model selection based on their needs (see Section~\ref{sec:fairness} for an illustrative example). In the individual setting, clients evaluate their Rashomon sets locally for further use, without revealing any information to the server.

\section{Empirical Study}\label{sec:empiricalstudy}
To illustrate how the proposed definitions operate in practice and how they compare across settings, we conduct an empirical study\footnote{\href{https://github.com/xheilmann/FederatedLearningMultiplicity}{https://github.com/xheilmann/FederatedLearningMultiplicity}} implementing the full pipeline described above, covering the three definitions of federated Rashomon sets and including a small application to fairness analysis. 
In building the pool of FL model candidates, we adopt in this study the re-training strategy, as it remains the standard baseline in current centralized Rashomon set research~\cite{hsu2024dropout, hsu2024rashomongb, ganesh2025systemizingmultiplicitycuriouscase}. The goal of this empirical study is to illustrate the behavior and implications of the proposed federated Rashomon set definitions across a range of controlled experimental settings. As this paper is primarily a theoretical contribution, the empirical section is intentionally designed to provide targeted illustrations of the introduced definitions. A systematic comparison of candidate generation methods, model architectures, FL settings, or alternative paradigms such as personalized FL would constitute a substantial study on its own, and we therefore leave such investigations to future work.

\subsection{Setup}

We conduct our empirical evaluation on three widely studied benchmark datasets in the FL literature. First, we use the Dutch Census dataset~\cite{Dutch}, where the task is to predict occupational status. Second, we employ the ACS Income dataset~\cite{adult}, using the simplified variant introduced by~\citet{corbucci2025benefits}, to predict whether annual income exceeds $\$50.000$. Lastly, we include the MNIST dataset~\cite{mnist}, a multi-class classification benchmark in which the goal is to identify the handwritten digit depicted in each image. The first two datasets are widely used in the FL fairness literature, enabling us to later analyze how our framework interacts with fairness considerations.
To generate the Rashomon set candidates, we adapt the re-training strategy to a federated setting. Here, as the underlying global model architecture, we use a single-layer classifier for Dutch and ACS Income and a two-layer neural network with a ReLU activation in the hidden layer for MNIST. As a baseline model, we train for each configuration a model for 100 training rounds on all clients. For the Rashomon set candidate models, we re-train 400 models for each configuration in a cross-silo FL setting, FedAVG as aggregation mechanism, and with simulated heterogeneous client distributions (see Appendix~\ref{app:training} for more details about communication and computational costs of training and Appendix~\ref{app:tuning} for details about the hyperparameter tuning). The re-training is performed by varying the size of the set of clients participating in each training round $|\psi|$, the number of rounds $R$, and the random initializations. After the training process has finished, we construct the Rashomon sets using accuracy as the performance metric underlying $\delta$, and consider $\epsilon \in \{0.0, 0.004, 0.008, 0.012, 0.016, 0.020, 0.024, 0.028, 0.032, 0.036, 0.040\}$. 
Additionally, for the Dutch dataset, we perform an ablation study varying the number of participating clients across five configurations: 10, 20, 30, 40, and 50 clients. For all other experiments, we fix the size to 20 participating clients. The aim of this is to understand if and how the Rashomon set construction and multiplicity metrics change when the number of FL clients increases.

We evaluate all candidate models under the three Rashomon set definitions introduced in Section~\ref{sec:characterizing}, using all clients as evaluation clients. For the $t$-agreement Rashomon sets, we consider $t \in \{0.6, 0.75, 0.9\}$. To evaluate the individual Rashomon set, we randomly sample 10 clients. For each Rashomon set, we compute all the metrics described in Section~\ref{sec:multimetrics} as well as the Rashomon ratio~\cite{semenova2022existence}, which quantifies the relative volume of the set within the hypothesis space. Small Rashomon ratios could imply a harder learning problem, while higher ones might imply simpler ones.

\begin{figure}
    \centering
    \includegraphics[width=\linewidth]{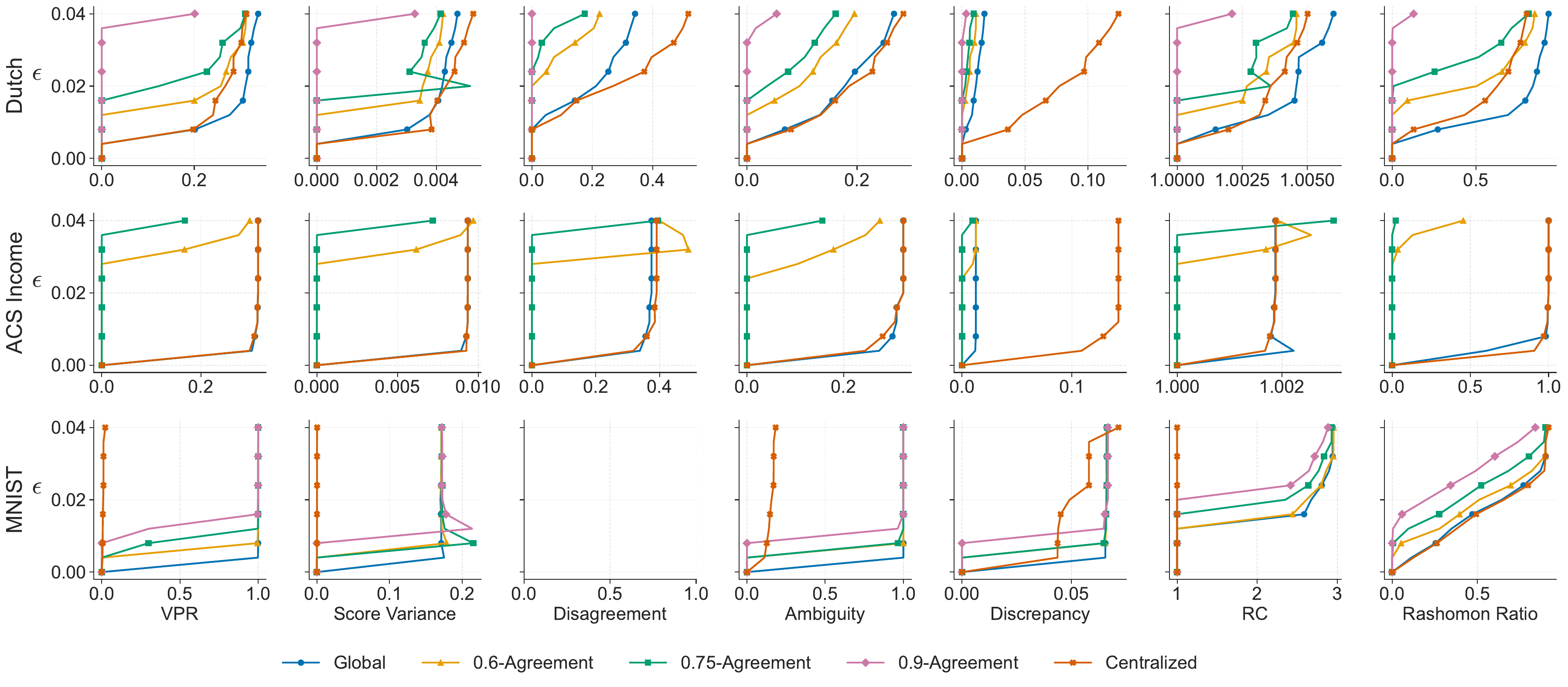}
    \caption{Comparison of multiplicity metrics on Rashomon sets defined using the $t$-agreement and global definition, with centralized evaluation as baseline for Dutch, ACS Income, and MNIST. Global definition yields consistently higher multiplicity, $t$-agreement sets are smaller, and centralized evaluation shows higher discrepancy. With MNIST, Rashomon ratios and multiplicity metrics are higher due to the complexity of the problem. Disagreement is only calculated for the binary outcomes. }
    \label{fig:global}
\end{figure}

As a baseline, we evaluate how the Rashomon sets and multiplicity metrics would have behaved in a centralized test setting. Concretely, after generating the Rashomon set candidates, we construct a centralized test set and use it to derive Rashomon sets based on the centralized measurements and compute all predictive multiplicity metrics thereof.

\begin{figure}
    \centering
    \includegraphics[width=\linewidth]{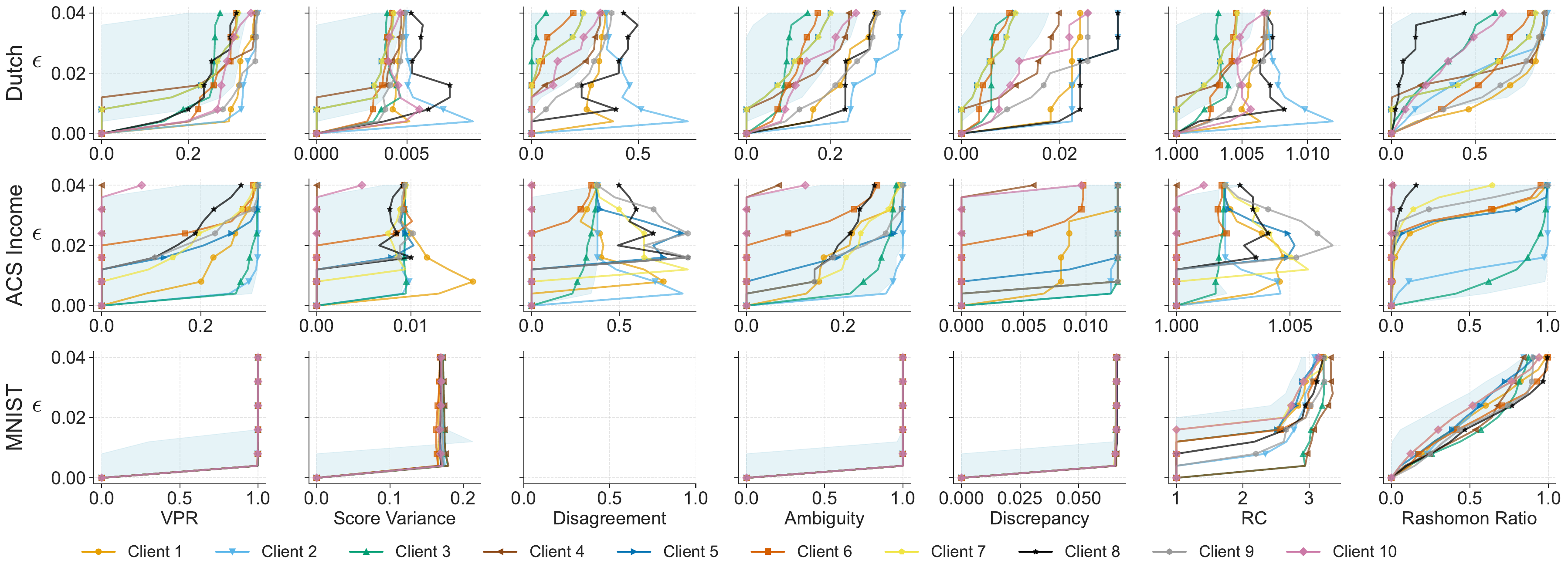}
    \caption{Comparison of multiplicity metrics for individual Rashomon sets (10 clients), with the blue shaded area showing the min-max range from Figure~\ref{fig:global}. Disagreement is only defined for binary outputs. Clients frequently deviate from the global range, indicating that it is essential to incorporate an individual Rashomon set definition for capturing local differences.}
    \label{fig:individual}
\end{figure}

To provide a valuable first comparison of all Rashomon set definitions, we assume a setting where clients trust the central server, and therefore, we report all score-based predictive multiplicity metrics as well as disagreement in the main paper without adding differential private noise. The results for all experiments with added Differential Privacy are reported in Appendix~\ref{app:diffpriv}. Specifically, our goal here is to formalize the Rashomon set definitions in FL and demonstrate that existing metrics can be adapted to this setting. Differential Privacy would obscure the underlying results, thus not serving this theoretical objective. Furthermore, as \textit{Disagreement} can only be evaluated in closed form for binary classifiers, we show results for this metric only for ACS Income and Dutch.

\subsection{Comparison of Definitions}
Figure~\ref{fig:global} shows the experimental results for the global and $t$-agreement Rashomon set definitions. We provide the cumulative distributions in Appendix~\ref{app:cum}. For the Dutch dataset, the global definition results in higher Rashomon ratios and higher multiplicity metrics. In contrast, the $t$-agreement definition yields smaller Rashomon sets, reflected in lower Rashomon ratios, and requires more permissive $\epsilon$ constraints to find valid sets. Yet, multiplicity metrics are lower. The centralized evaluation generally lies between or close to the global and the $t$-agreement definitions. Notably, the discrepancy metric is substantially higher under the centralized evaluation, suggesting that employing the FL Rashomon set definitions reduces prediction volatility between models in the Rashomon sets compared to building these based on centralized test data.  

For the ACS Income dataset, we observe similar results. However, for the 0.9-agreement setting, no Rashomon sets are found within the predefined constraint range. Overall, the $t$-agreement strategy identifies considerably fewer models under tight performance constraints. Centralized evaluation aligns with global evaluation across most metrics, except discrepancy, which again reaches exceptionally high values in the centralized evaluation. 

For MNIST, a 10-class classification task, both the Rashomon ratios and the multiplicity metrics are higher than in the previous binary classification tasks, and the centralized baseline differs significantly from the FL settings. This may reflect the increased complexity of the problem. 

\begin{figure}
    \centering
    \includegraphics[width=\linewidth]{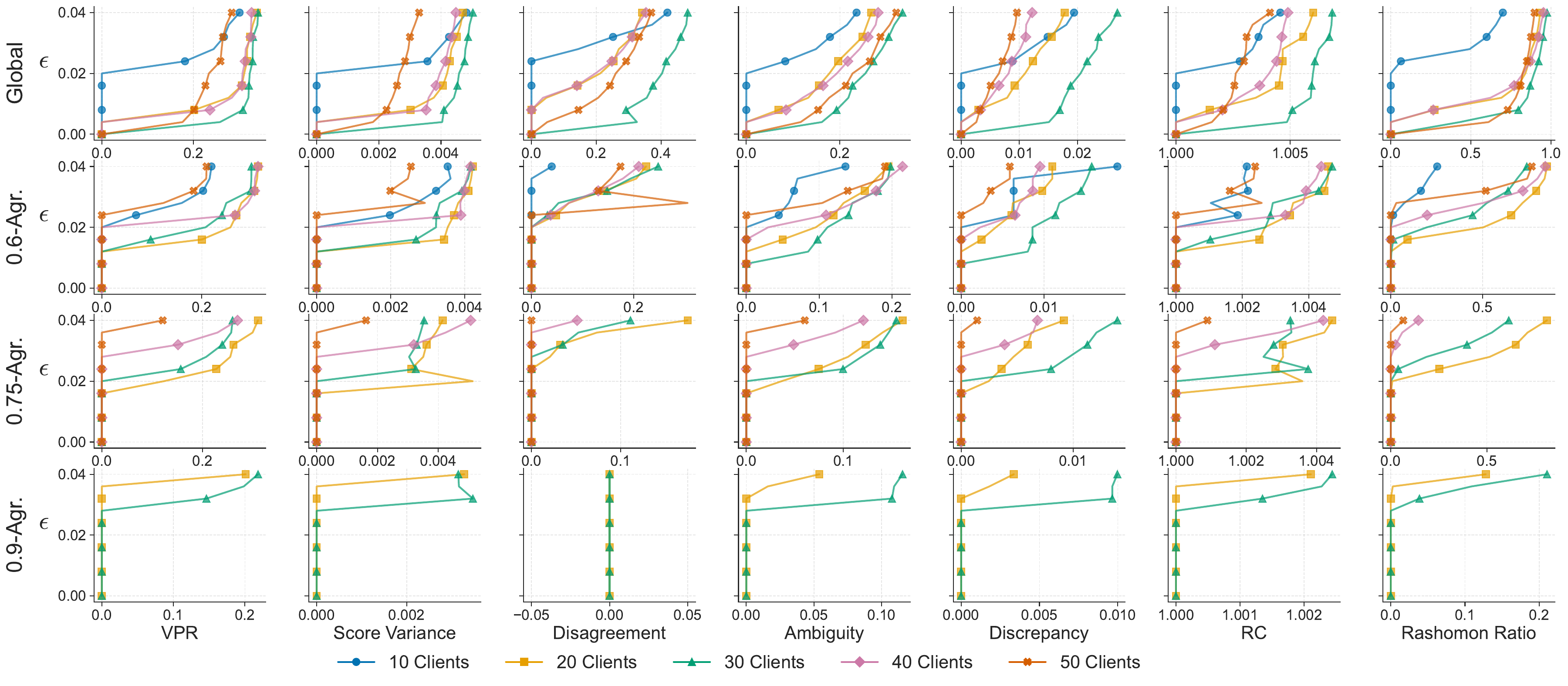}
    \caption{Multiplicity metrics for global and $t$-agreement Rashomon sets on the Dutch dataset when varying the number of FL clients. The metrics remain consistent across different clients, indicating that the analysis scales. Stricter $t$-agreement thresholds (e.g., $0.9$) fail to produce Rashomon sets for 10, 40, or 50 clients; tight constraints can limit the feasibility in extreme client configurations.}
    \label{fig:client_numbers}
\end{figure}
Figure~\ref{fig:individual} presents results for the individual Rashomon set definition. To contextualize these results with the global and $t$-agreement definitions, we overlay a blue-shaded region between the maximal and minimal value observed in Figure~\ref{fig:global}. While some individual clients fall within this range, others behave differently, with a Rashomon capacity and disagreement value that is generally higher at the individual level. For ACS Income, most of the metrics fall within the shaded area, which may be due to the higher number of datapoints per client. 
Consequently, neither global aggregation nor the $t$‑agreement definition fully captures the heterogeneity of model behavior on individual clients. Incorporating the individual Rashomon set definition is therefore essential to capture local differences and client-specific outliers.
Moreover, this approach allows clients to define their own performance constraints when constructing Rashomon sets.

Lastly, Figure~\ref{fig:client_numbers} presents the results for the Dutch dataset across varying numbers of clients participating in the FL training (10, 20, 30, 40, 50). This experiment was designed to evaluate the scalability of our pipeline as the number of clients increases. Overall, the results remain consistent, without extreme outliers, demonstrating that our approach scales effectively across different federation sizes. However, we observe that under small (10 clients) and large (40–50 clients) federation sizes, the $t$-agreement Rashomon set definition struggles to identify models within tight $\epsilon$ constraints as $t$ increases. Notably, no Rashomon sets are found for the $t=0.9$ agreement setting in these configurations. Globally, all multiplicity metrics follow consistent trends across varying client numbers. 
These results show that, aside from a very strict $t$-agreement threshold, the proposed definitions are robust and scalable with respect to the federation size.

\subsection{Application to Fairness}\label{sec:fairness}
We now illustrate how Rashomon sets enable FL clients to select the models that best align with their local preferences, rather than relying on a single ``best'' model based on aggregate performance. This approach allows for personalized model selection based on secondary objectives while maintaining comparable global predictive performance across all models in the Rashomon set. In our demonstration, we focus on a fairness application and specifically Demographic Parity~\cite{barocas2017nips} as a fairness criterion. This analysis is restricted to the Dutch and the ACS Income dataset, as they contain demographic information that can be used to measure model fairness.

\begin{figure}
    \centering
    \includegraphics[width=0.97\linewidth]{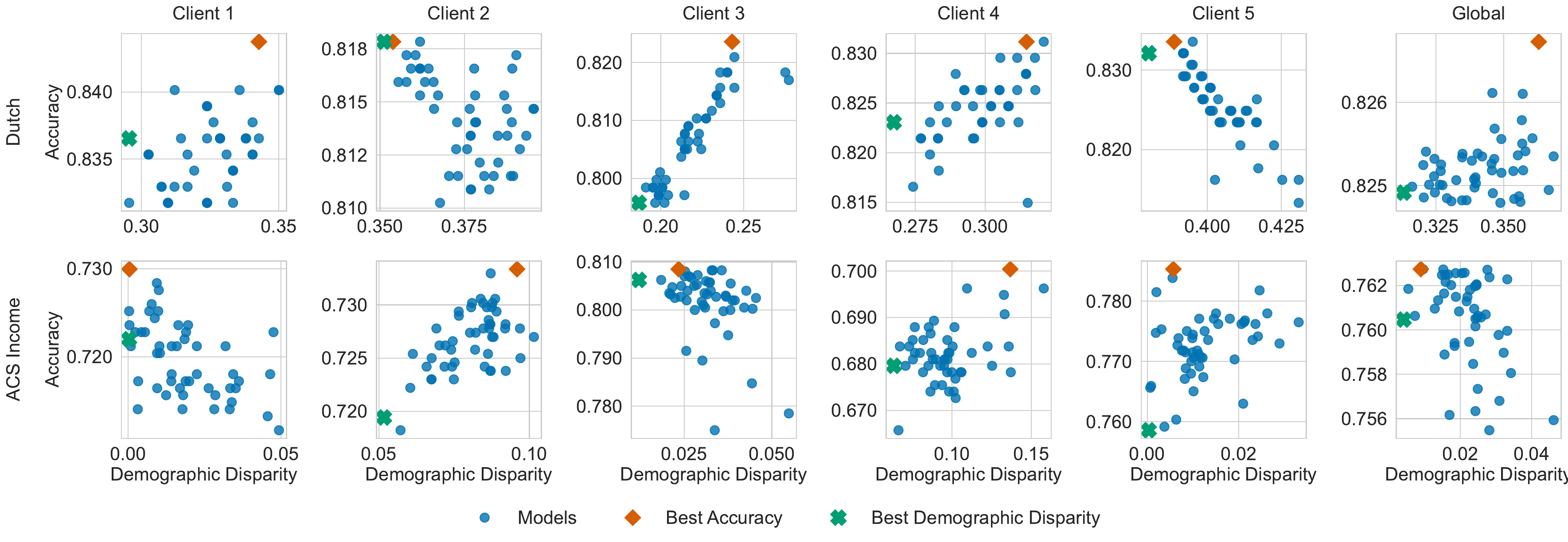}
    \caption{Demographic Disparity (on the x-axis, the lower the better) and accuracy values (on the y-axis, the higher the better) for 50 models in the epsilon smallest global Rashomon set shown for 5 clients on the Dutch and ACS Income data. }
    \label{fig:deno_par}
\end{figure}

For this demonstration, we evaluate 50 models sampled from the global Rashomon sets for each dataset on each client's local data. We report both accuracy and Demographic Disparity (DD), defined as the maximum difference between the Demographic Parity of the different sensitive groups $DD= \max_{y\in Y} \max_{z\in Z} \mathbb P(\hat{Y}=y \mid Z=z) - \mathbb P(\hat{Y}=y \mid Z \neq z)$ where $y$ is a predicted target and $z$ a sensitive attribute. In simple terms, predictions $\hat{Y}$ should not depend on $Z$.
For both Dutch and ACS Income, DD is computed with respect to \textit{sex}, which we treat as the sensitive attribute. Figure~\ref{fig:deno_par} presents results for five clients on the $\epsilon$ smallest global Rashomon set (Dutch: $\epsilon=0.008$, ACS Income: $\epsilon = 0.004$) against the global evaluation (results for all other $\epsilon$ values can be found in Appendix~\ref{app:fairness}).

The spread of these models varies across clients, reflecting heterogeneity in local data distribution. For both datasets, DD values differ significantly across clients and often differ from the global evaluation (last column), emphasizing the importance of a client-specific evaluation step in FL pipelines~\cite{corbucci2025benefits}. For the Dutch dataset, however, models achieving the best DD value (the lower the better) generally maintain accuracy close to that of the globally highest-performing model, except client 3, where the fairest model exhibits the lowest accuracy. For the ACS Income dataset, client 2 and client 4 show the highest accuracy-fairness trade-off. Overall, this application underscores the value of exploring Rashomon sets in FL: practitioners are assured that all models in the Rashomon set maintain high global performance, while retaining the flexibility to select those that best satisfy local data, requirements, and objectives.

\section{Conclusion}\label{sec:conclusion}
In this paper, we establish the foundation for defining Rashomon sets and evaluating predictive multiplicity metrics in FL settings. We introduce three federated Rashomon set formulations, namely global, $t$-agreement, and individual, and show how multiplicity metrics can be computed for each of them within an FL scenario. Beyond the theoretical contribution, we conduct an empirical study demonstrating both the validity of these definitions and their practical relevance. Our findings suggest that Rashomon set analysis enables practitioners to move beyond the conventional ``single best model'' approach used in the classic FL pipelines, supporting more informed and robust model selection.

Building on these foundations, a direction for future work is the development of alternative strategies for constructing the Rashomon set candidates that leverage the inner dynamics of FL, such as intermediate model updates or aggregation mechanisms, rather than relying only on re-training. A comprehensive empirical evaluation of such strategies across diverse FL settings, along with a systematic comparison of the proposed formulations under different FL paradigms, would provide deeper insight into their practical behavior.

Moreover, while our analysis focuses on FL as a decentralized, multi-party scenario, other secure multi-party frameworks~\cite{SMPC-survey}, such as CryptTen~\cite{crypten}, introduce additional heterogeneity that may require extending our framework, representing a promising direction for future work. 

%Finally, integrating governance and auditing mechanisms into multiplicity-aware FL frameworks remains an important open challenge. This includes designing methods to ensure accountability across clients and to support compliance with regulatory requirements.

\clearpage

\section*{Acknowledgment}
XH and MC were supported by the ``TOPML: Trading Off Non-Functional Properties of Machine Learning'' project funded by Carl Zeiss Foundation, grant number P2021-02-014. 

\section*{Generative AI usage statement}

LLMs were used to aid non-native speakers with grammar and word corrections. During the implementation of the code needed for this submission, we used LLMs for autocompletion of code and to fix bugs.

%\clearpage

\bibliographystyle{ACM-Reference-Format}
\bibliography{ref}
\clearpage
\appendix

\section{Table of Notation}\label{app:symbols}

We report in Table~\ref{tab:symbols} a list of the symbols used in the paper to increase clarity and readability.

\begin{table}[htbp]
    \centering
    \begin{tabular}{l l | l l}
        \toprule
        \textbf{Symbol} & \textbf{Description} & \textbf{Symbol} & \textbf{Description} \\
        \midrule
        $\mathcal{C}$ & set of clients & $\Delta$ & set of metric functions \\
        $\mathcal{C}_E$ & set of evaluation clients & $f_E$ &evaluation aggregation algorithm\\ 
         $C$ & number of clients & $\mathcal{E}$ & threshold set\\
        $h$ & model & $\epsilon_i$ & threshold  \\
        $\hat{h}$ & local model & $P_\calR$ &distribution of models in $\calR$\\
        
        $R$ & FL rounds & $\delta_i$ & metric function \\
        $S$ & server & $h_{\bw^*}$ & baseline Model \\
        $r_i$ & each round ($r_i \in R$) & $m$ & Rashomon set size \\
        $\psi$ & subset of selected clients & $\mathcal{H}$ & set of all models \\
        $E$ & local epochs & $\mathcal{R}^m$ & empirical Rashomon set of size $m$\\
        $b_i$ & batch $i$ & $(\delta^M, \epsilon^M)$ & Multiplicity constraint \\
        $\hat{g}_c$ & local gradient of client $c$ & $n$ & \# samples  \\
        $s, \bs$ & score, score vector (decision-based metric) & $n_c$ & \# samples of client $c$ \\
        $\tau$ & threshold (decision-based) & $\calD$ & Dataset \\
        $\gamma$ & Discrepancy & $\alpha$ & Ambiguity\\
        $\mu$ & Disagreement & $t$& agreement parameter \\
        $rc$ & Rashomon capacity & $v$ & viable prediction range\\
        $sd$& standard deviation &\\
        \bottomrule
    \end{tabular}
    \caption{Table of notation.}
    \label{tab:symbols}
\end{table}

\section{Differential Privacy}\label{app:diffpriv_explanation}
Differential Privacy is a privacy-enhancing technology~\cite{10.1007/11787006_1} that ensures that running the same algorithm $\mathcal{M}$ on two neighbouring datasets produces indistinguishable outputs up to an upper bound $\varepsilon$ called the privacy budget. More formally, 
\begin{definition}
\label{def:dp}
\textit{An algorithm $\mathcal{M}$ satisfies ($\varepsilon$)-Differential Privacy, where $\varepsilon > 0$, if for any pair of ``neighbouring'' datasets $\calD$, $\calD'$ differing in exactly one entry, and for any set of outputs $\mathcal{O}$ the following condition holds: $$P[\mathcal{M}(\calD) \in \mathcal{O}] \leq e^{\varepsilon} P[\mathcal{M}(\calD') \in \mathcal{O}] $$}
\end{definition}
In this paper, we apply Differential Privacy to share Differentially Private cumulative distributions that are shared with the server by the clients. 
We implemented Differential Privacy in our pipeline using the diffprivlib~\cite{diffprivlib} library. Specifically, we used the ``tools.histogram'' function that computes a differentially private histogram of a set of data, adding noise using a Geometric Truncated mechanism. In our experiments, we fixed $\varepsilon=0.1$.

\section{Explanations of Score-based Predictive Multiplicity Metrics}\label{app:metrics}
\subsubsection{Rashomon Capacity}

The Rashomon capacity (RC) was proposed in~\citet[Definition~2]{hsu2022_rashomoncapacity} and measures the spread of output scores for $d$-class classification problems in the probability simplex $\Delta_s$. The mathematical formulation is 
\begin{equation}
    rc(\bx_i)\triangleq \sup\limits_{P_{\calR}} \inf\limits_{\bq \in \Delta_s} \E_{h \sim P_{\calR}} D_\textsf{KL}(h(\bx_i)\|\bq),
\end{equation}
where $P_{\calR}$ is the probability distribution over the models in the Rashomon set, and $D_\textsf{KL}(\cdot\|\cdot)$ is the Kullback-Leibler (KL) divergence. The infimum $\inf\limits_{\bq \in \Delta_s} \E_{h \sim P_{\calR}} D_\textsf{KL}(h(\bx_i)\|\bq)$ captures, in KL divergence terms, how the prediction scores for a sample $\bx_i$ given a distribution $P_\calR$ over all the models $h$ in the Rashomon set are spread out. The $q$ acts as a ``centroid'' for the outputs of the classifiers.
Taking the supremum over all $P_\calR$ selects the worst-case distribution in the Rashomon set.
Thus, the Rashomon Capacity can be seen as an instance of the information diameter, a measure of the ``size’’ of a family of probability distributions~\citep{kemperman1974shannon}. 

\subsubsection{Viable Prediction Range and Standard Deviation}
Viable Prediction Range (VPR)~\cite{watson2023predictive} calculates the largest score deviation of a sample that can be achieved by models in the Rashomon set:
\begin{equation}
    v(\bx_i)\triangleq \max\limits_{h \in \calR} h(\bx_i) - \min\limits_{h \in \calR} h(\bx_i).
\end{equation}
Instead of this, one can also measure the standard deviation of the scores of a sample by all models in the Rashomon set~\cite{long2023arbitrariness}:
\begin{equation}
    sd(\bx_i) \triangleq \sqrt{\E_{h\sim P_\calR}[(h(\bx_i) - \E_{h\sim P_\calR}[ h(\bx_i)])^2] }.
\end{equation}

\section{Training of the Models}\label{app:training}
We conducted all experiments on a machine equipped with a 256 AMD EPYC 9554 64-Core Processor and 1TB of RAM. All experiments were conducted solely on CPUs, as no dedicated GPU was available on this system.
Table~\ref{tab:performance_metrics} reports the average time needed to train a single model and the total time required to train the 400 Rashomon set candidate models used to construct the Rashomon sets for the Dutch, ACS Income, and MNIST datasets.
As described in the main text, we used a single-layer classifier for Dutch and ACS Income, and a two-layer neural network with a ReLU activation in the hidden layer for MNIST. 
These relatively simple architectures were selected to make the re-training approach feasible within an FL context. Furthermore, we performed re-training by varying the size of the set of clients participating in each training round $|\psi|$, the number of rounds $R$, and the random initializations. Specifically, we choose $|\psi| \in \{\frac{|\mathcal{C}|}{4}, \frac{|\mathcal{C}|}{2}, \frac{3|\mathcal{C}|}{4}, |\mathcal{C}|\}$ for client set $\mathcal{C}$ and $R \in [1,10]$.
Re-training remains the standard approach for generating Rashomon set candidates in the centralized learning literature, even though it is resource-intensive. 
Despite this, we were able to obtain the Rashomon set with 400 models in a reasonable time, as shown in Table~\ref{tab:performance_metrics}. 
We acknowledge, however, that applying this technique to larger or more complex models becomes impractical.

With regard to communication costs, the complete pipeline introduced in Section~\ref{sec:pipeline} adds at most two communication rounds on top of the standard FL setting.
An important direction for future work is to develop alternative strategies that reduce training time by exploiting the internal workings of FL.

\begin{table}[h]
    \centering
    \begin{tabular}{l c c c} 
        \toprule
        \textbf{Dataset} & \textbf{Clients $|\mathcal{C}|$} & \textbf{Avg. Training Time (s)} & \textbf{Total Training Time (s)} \\
        \midrule 
        ACS Income & 20  &  $15.16\pm0.10$ &  2h 40m 42s\\
        MNIST & 20 & $82.10\pm0.68$  & 10h 1m 41s\\
        Dutch & 10 & $17.00\pm0.11$ & 2h 49m 44s\\
        Dutch & 20 & $16.49\pm0.13$ & 2h 48m 21s\\
        Dutch & 30 & $13.68\pm0.16$ & 2h 24m 47s \\
        Dutch & 40 & $16.29\pm0.11$ & 2h 42m 52s\\
        Dutch & 50 & $16.35\pm0.10$ & 2h 43m 20s \\
        \bottomrule 
    \end{tabular}
    \caption{Training time metrics across different datasets and client number}
    \label{tab:performance_metrics}
\end{table}

\section{Hyperparameter tuning}\label{app:tuning}

The hyperparameters used to train the models in the Rashomon sets were selected through a hyperparameter search. We used the Weights \& Biases~\cite{wandb} sweep feature, using Bayesian optimization to explore the search space.
Before starting the tuning phase, each client’s dataset was split into a training and a test set to ensure that the test data remained completely untouched during hyperparameter optimization. For each hyperparameter configuration, the training portion was further divided into a training and a validation set. This split was controlled by a random seed that changed for every hyperparameter test, reducing the risk of overfitting to a particular validation split.
Once the best hyperparameters, i.e., those maximizing validation accuracy, were identified, we retrained the models on the full training set.
The hyperparameters that we tuned are: batch size, learning rate, momentum, type of optimizer, and number of local epochs.

\section{Results with Differential Privacy}\label{app:diffpriv}
\begin{figure}
    \centering
    \includegraphics[width=\linewidth]{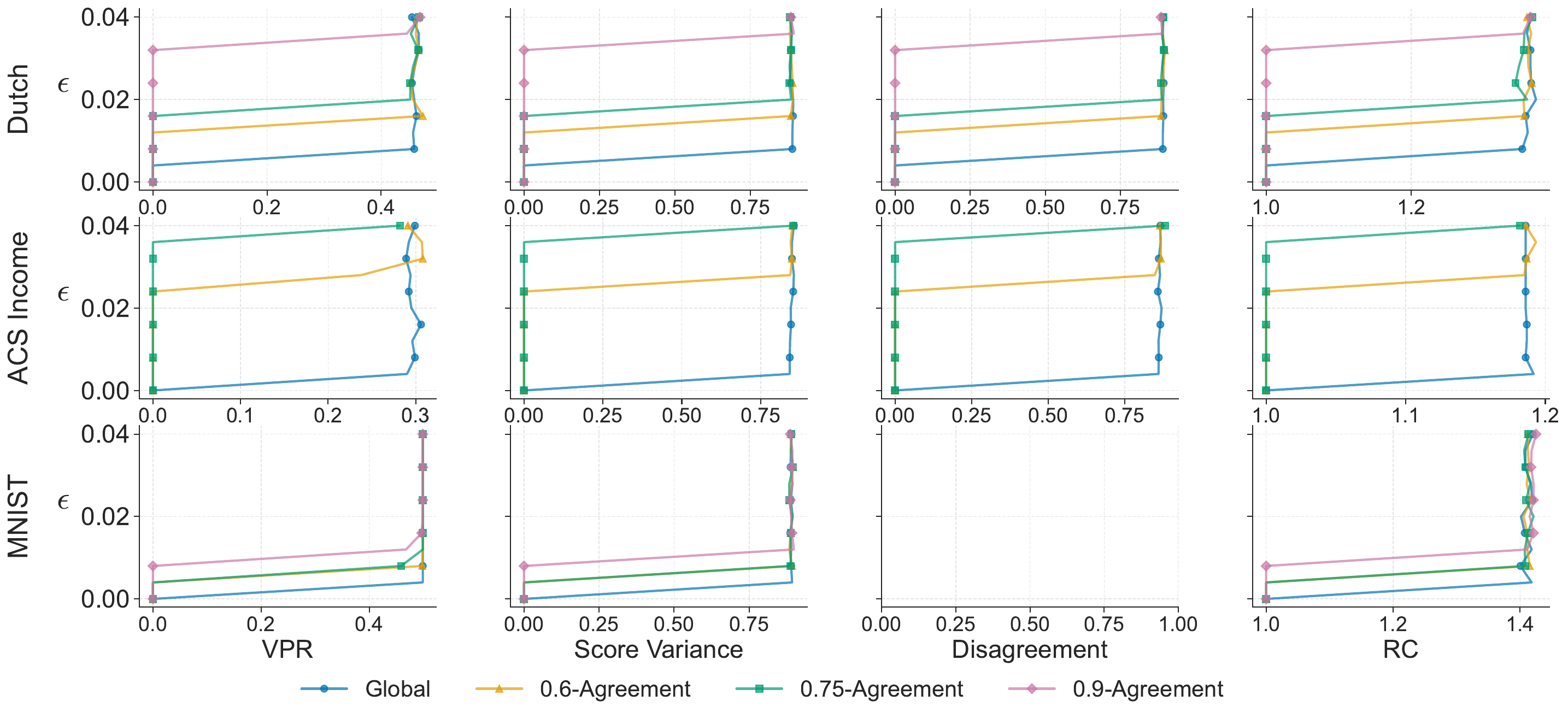}
    \caption{Comparison of multiplicity metrics defined on each sample with added Differential Privacy on Rashomon sets found with the $t$-agreement and global definition for Dutch, ACS Income, and MNIST.}
    \label{fig:global_dp}
\end{figure}

\begin{figure}
    \centering
    \includegraphics[width=\linewidth]{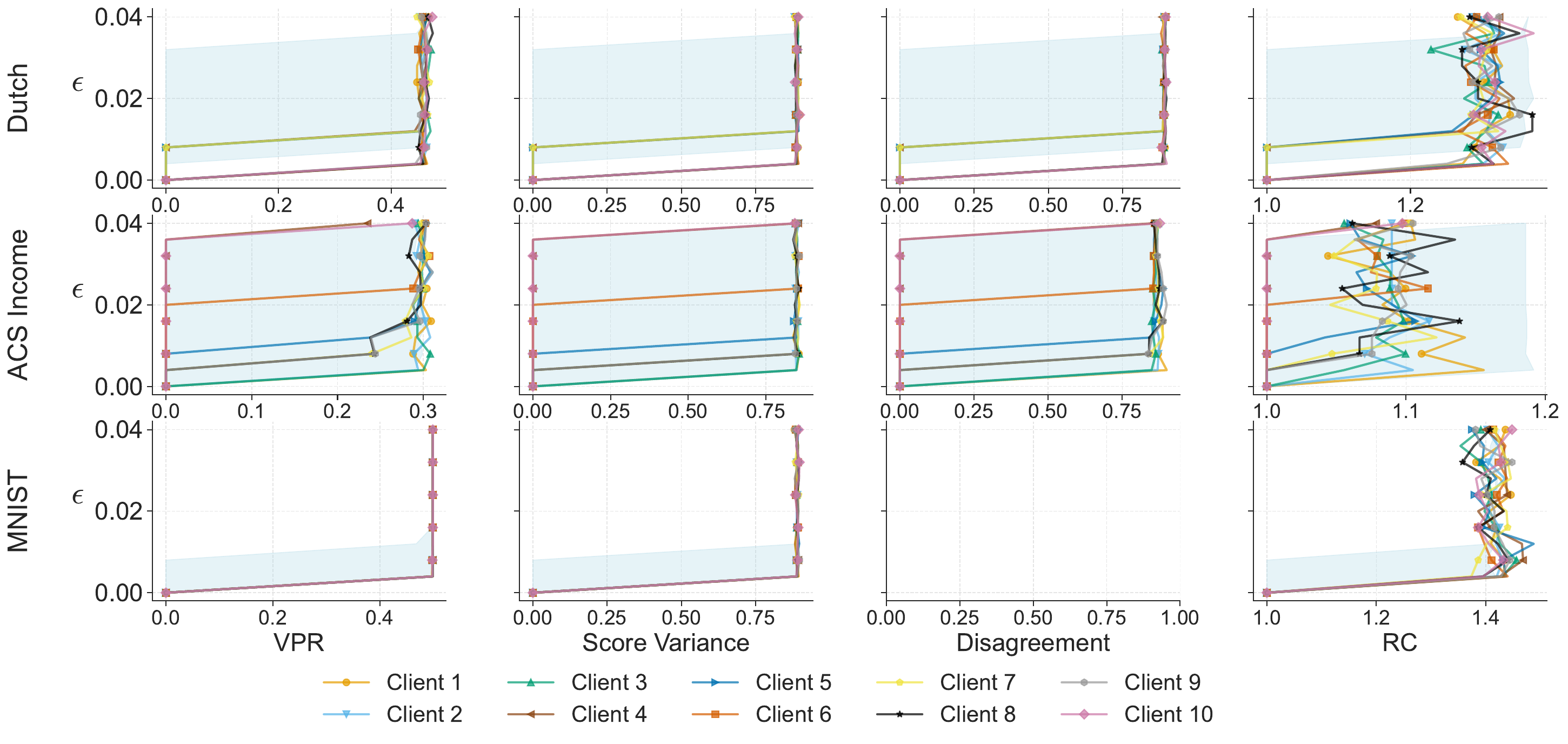}
    \caption{Comparison of multiplicity metrics defined on each sample with added Differential Privacy on Rashomon sets found with the individual definition for Dutch, ACS Income, and MNIST. We show results for 10 individuals.}
    \label{fig:individuals_dp}
\end{figure}

In Figure~\ref{fig:global_dp} and~\ref{fig:individuals_dp}, we provide results on the multiplicity metrics defined on each sample with added Differential Privacy~\cite{10.1007/11787006_1} as explained in Section~\ref{sec:multimetrics}. Results are based on 1000 buckets with a guarantee of $(\varepsilon=0.1)$-Differential Privacy. As mentioned in Section~\ref{sec:multimetrics}, the sparsity of the histogram bins leads to difficulties in evaluation. Especially considering the score variance in the non-private evaluation scores is extremely low, values with differential privacy do not show this trend at all. We therefore stress that additional work is needed here to make a private pipeline more robust against these difficulties.  

\section{Cumulative Distributions}
\label{app:cum}
\begin{figure}
    \centering
    \includegraphics[width=\linewidth]{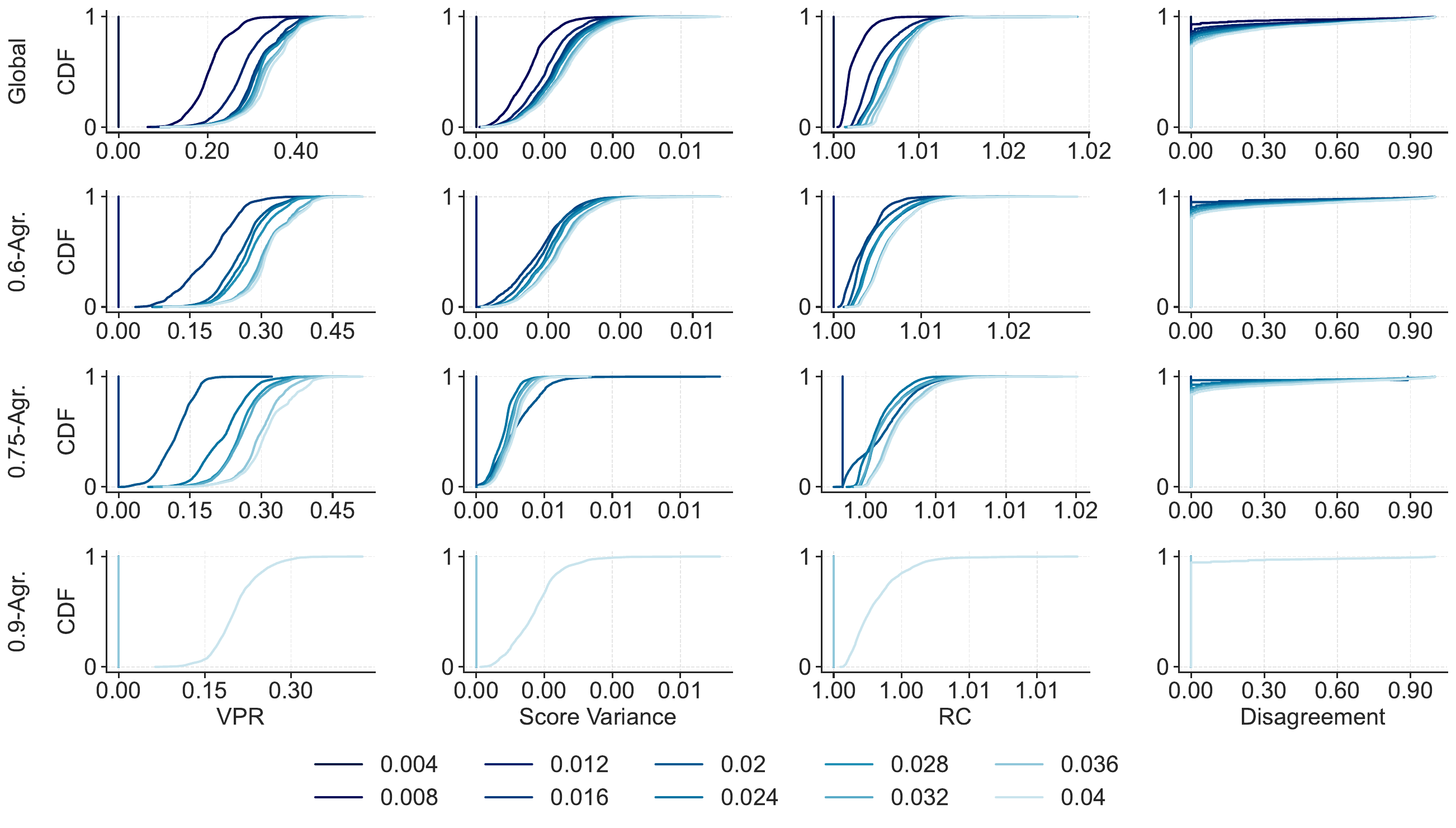}
    \caption{Cumulative distribution for each $\epsilon$ value over the score-based metrics and Disagreement for the Dutch dataset and 20 clients. Results are shown for the $t$-agreement and global definition.}
    \label{fig:dutch-cum}
\end{figure}

\begin{figure}
    \centering
    \includegraphics[width=\linewidth]{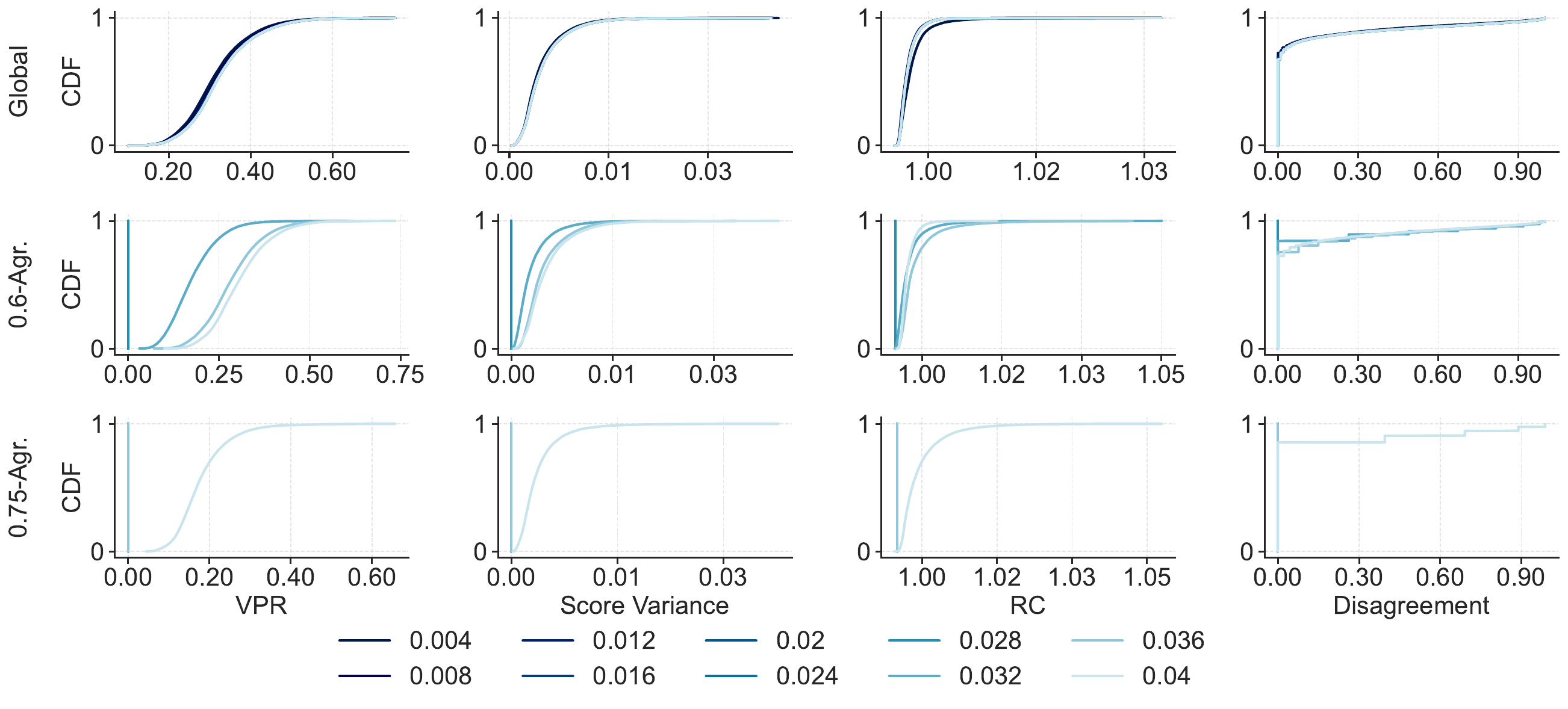}
    \caption{Cumulative distribution for each $\epsilon$ value over the score-based metrics and Disagreement for the ACS Income dataset. Results are shown for the $t$-agreement and global definition.}
    \label{fig:income-cum}
\end{figure}

\begin{figure}
    \centering
    \includegraphics[width=\linewidth]{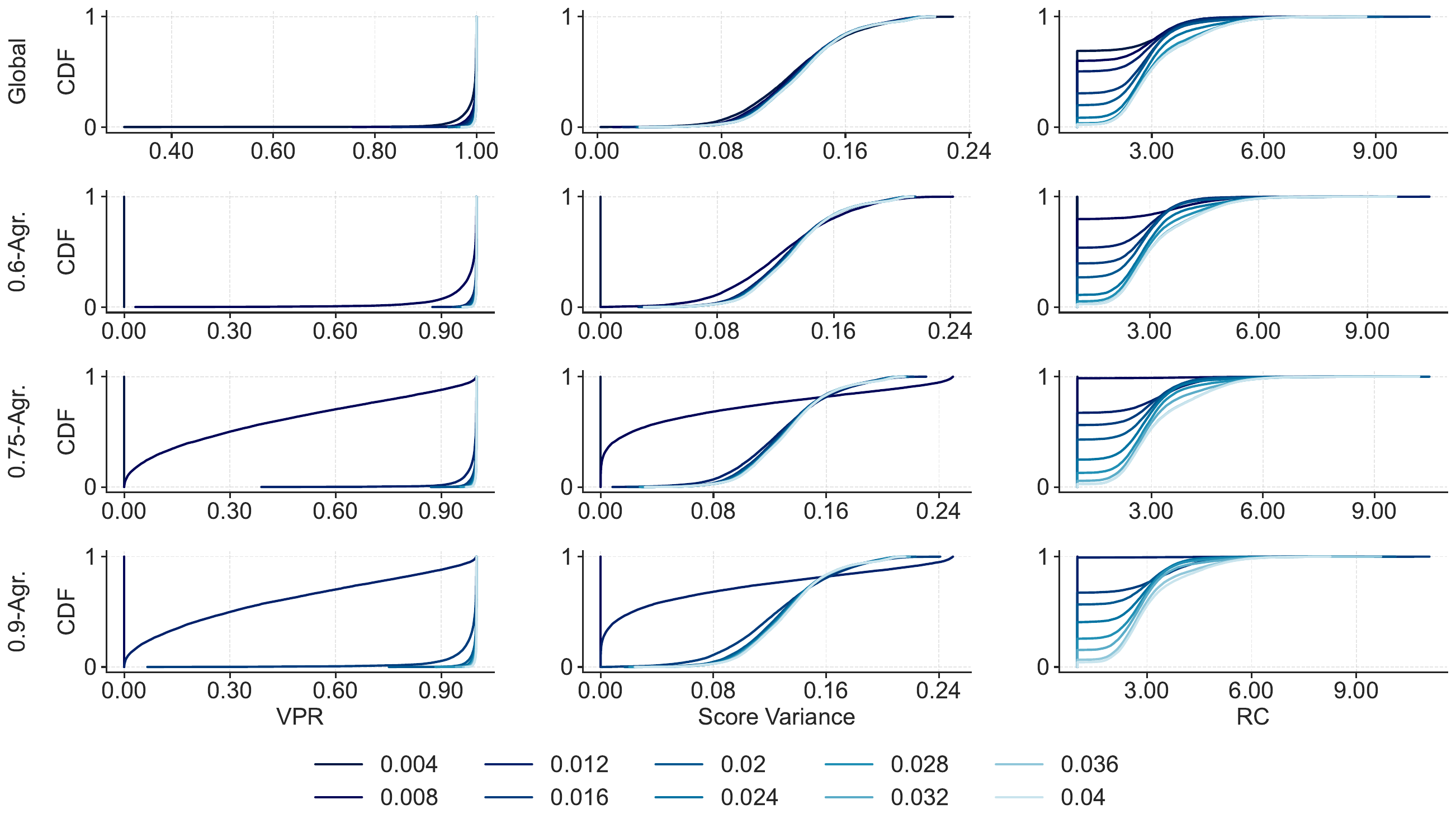}
    \caption{Cumulative distribution for each $\epsilon$ value over the score-based metrics for the MNIST dataset. Results are shown for the $t$-agreement and global definition.}
    \label{fig:mnist-cum}
\end{figure}

Figure~\ref{fig:dutch-cum},~\ref{fig:income-cum}, and~\ref{fig:mnist-cum} show the cumulative distribution of the multiplicity metrics, which are defined on each sample for the Dutch, ACS Income, and MNIST datasets. These plots show how the metrics are distributed across the complete dataset across all evaluating clients. For the figures in the paper, we took the percentiles from these underlying cumulative distributions. 


\section{Further Fairness Experiments}\label{app:fairness}
\begin{figure}
    \centering
    \includegraphics[width=\linewidth]{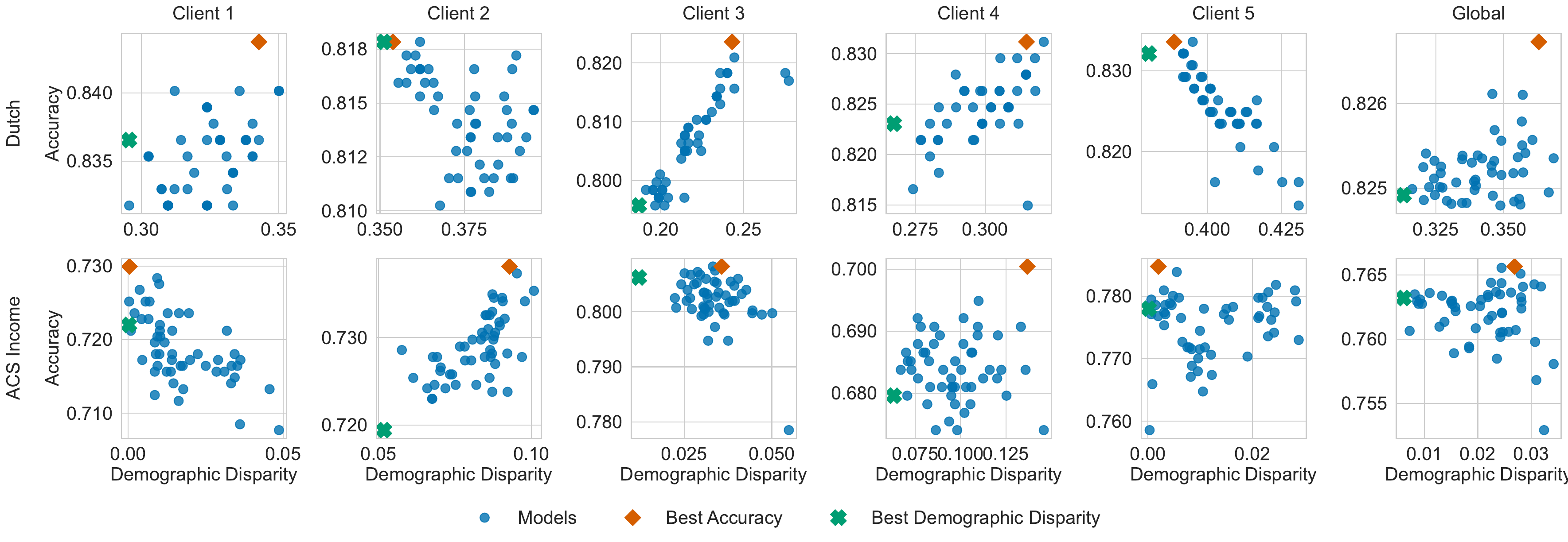}
    \caption{Demographic Disparity (on the x-axis, the lower the better) and accuracy values (on the y-axis, the higher the better) for 50 models in the $\epsilon=0.008$ global Rashomon set shown for 5 clients on the Dutch and ACS Income data.} 
    \label{fig:deno_par_008}
\end{figure}

\begin{figure}
    \centering
    \includegraphics[width=\linewidth]{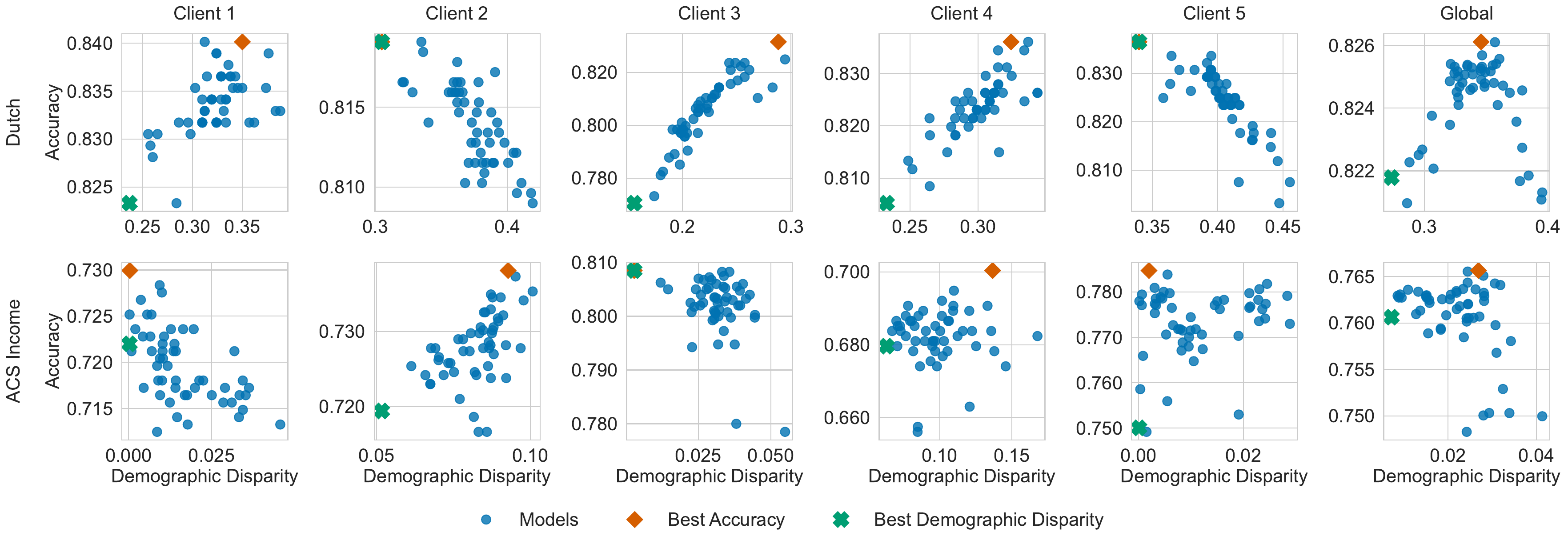}
    \caption{Demographic Disparity (on the x-axis, the lower the better) and accuracy values (on the y-axis, the higher the better) for 50 models in the $\epsilon=0.012$ global Rashomon set shown for 5 clients on the Dutch and ACS Income data.} 
    \label{fig:deno_par_012}
\end{figure}

\begin{figure}
    \centering
    \includegraphics[width=\linewidth]{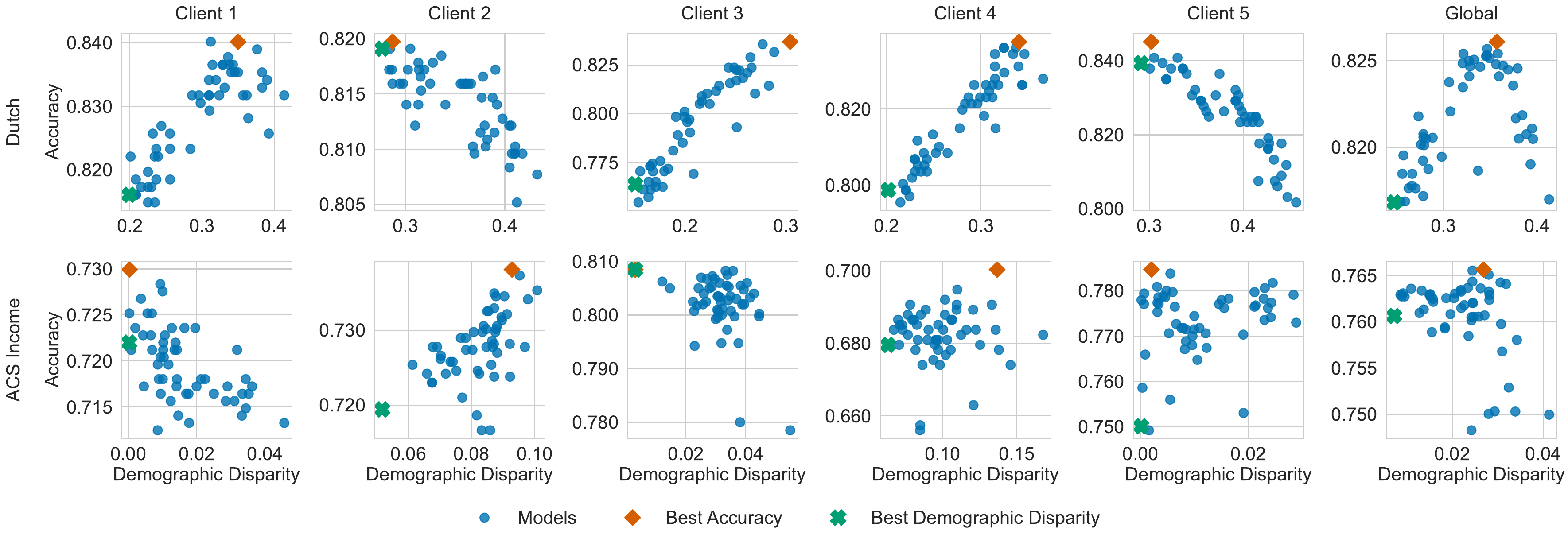}
    \caption{Demographic Disparity (on the x-axis, the lower the better) and accuracy values (on the y-axis, the higher the better) for 50 models in the $\epsilon=0.016$ global Rashomon set shown for 5 clients on the Dutch and ACS Income data.} 
    \label{fig:deno_par_016}
\end{figure}

\begin{figure}
    \centering
    \includegraphics[width=\linewidth]{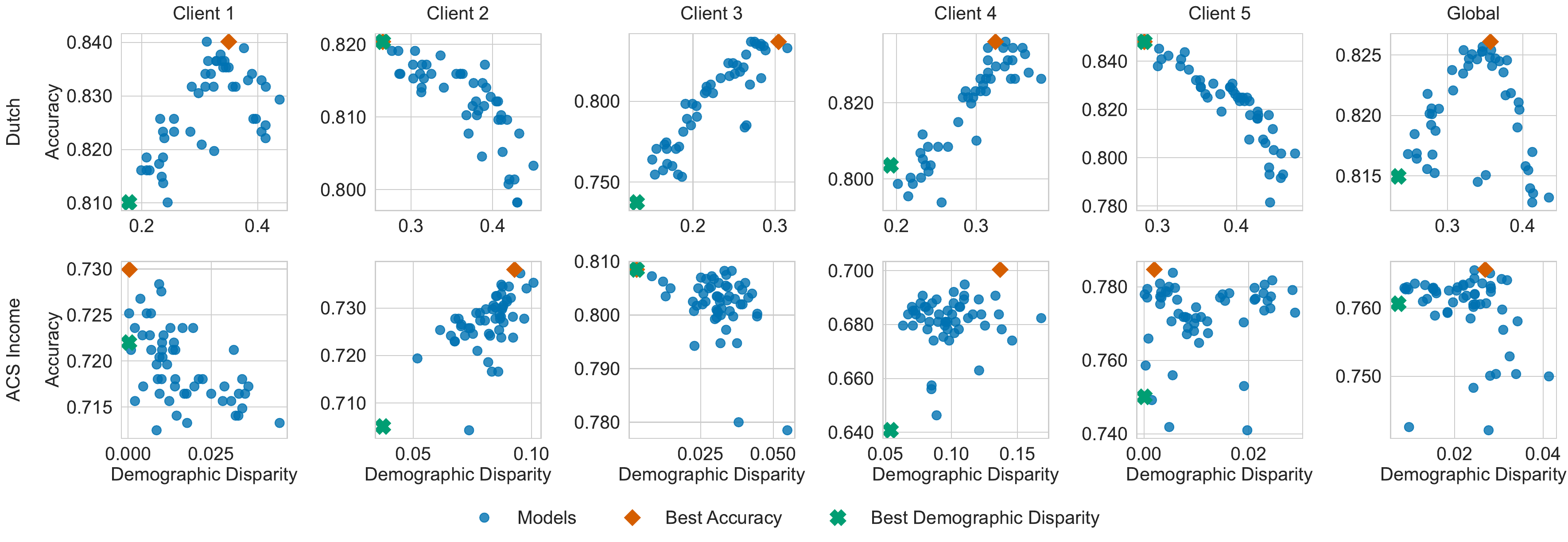}
    \caption{Demographic Disparity (on the x-axis, the lower the better) and accuracy values (on the y-axis, the higher the better) for 50 models in the $\epsilon=0.020$ global Rashomon set shown for 5 clients on the Dutch and ACS Income data.}
    \label{fig:deno_par_020}
\end{figure}

\begin{figure}
    \centering
    \includegraphics[width=\linewidth]{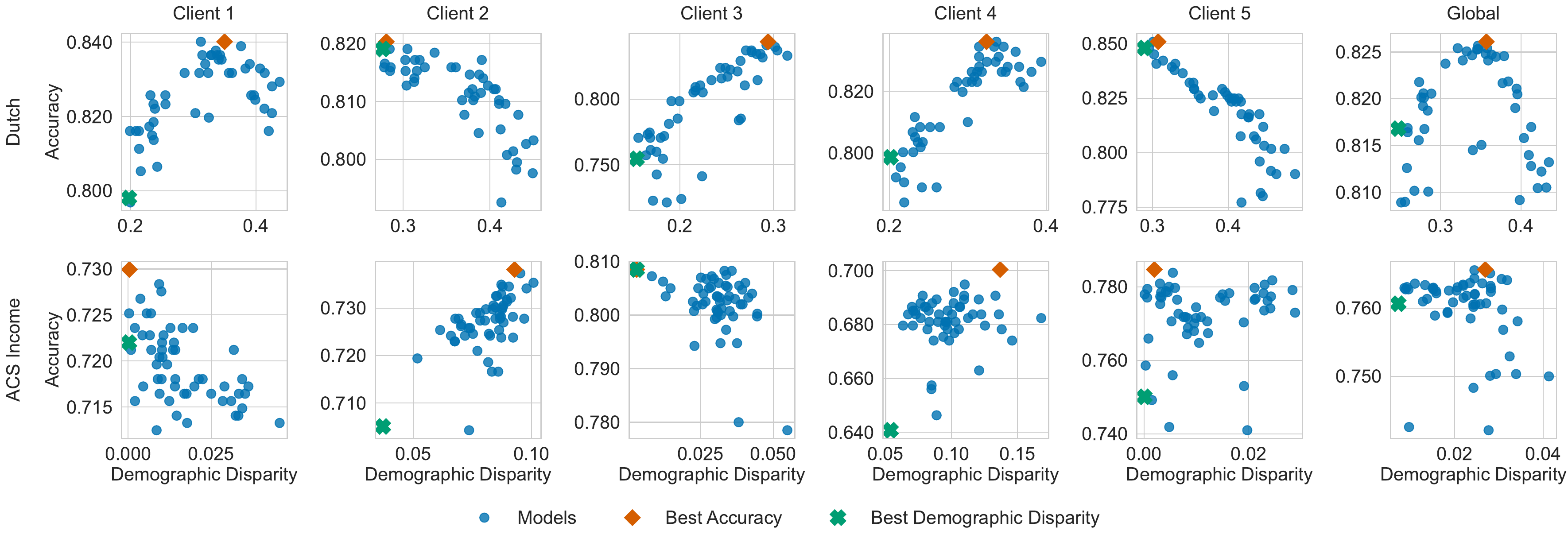}
    \caption{Demographic Disparity (on the x-axis, the lower the better) and accuracy values (on the y-axis, the higher the better) for 50 models in the $\epsilon=0.024$ global Rashomon set shown for 5 clients on the Dutch and ACS Income data.} 
    \label{fig:deno_par_024}
\end{figure}

\begin{figure}
    \centering
    \includegraphics[width=\linewidth]{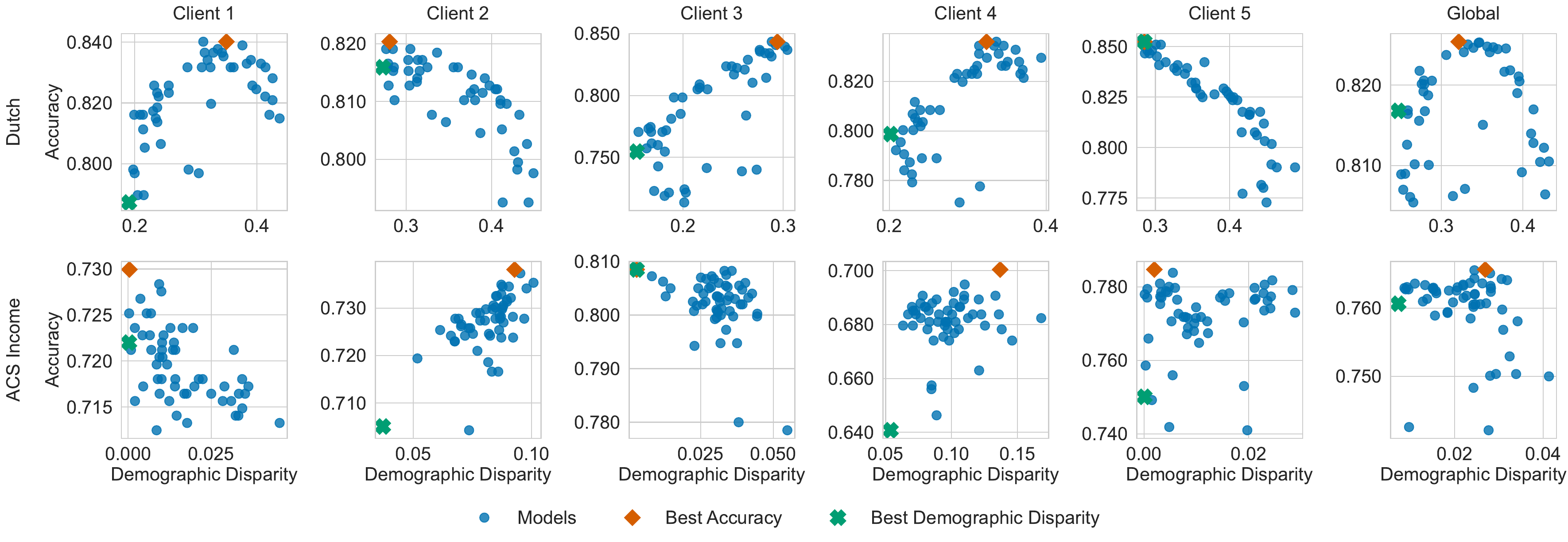}
    \caption{Demographic Disparity (on the x-axis, the lower the better) and accuracy values (on the y-axis, the higher the better) for 50 models in the $\epsilon=0.028$ global Rashomon set shown for 5 clients on the Dutch and ACS Income data.}
    \label{fig:deno_par_028}
\end{figure}

\begin{figure}
    \centering
    \includegraphics[width=\linewidth]{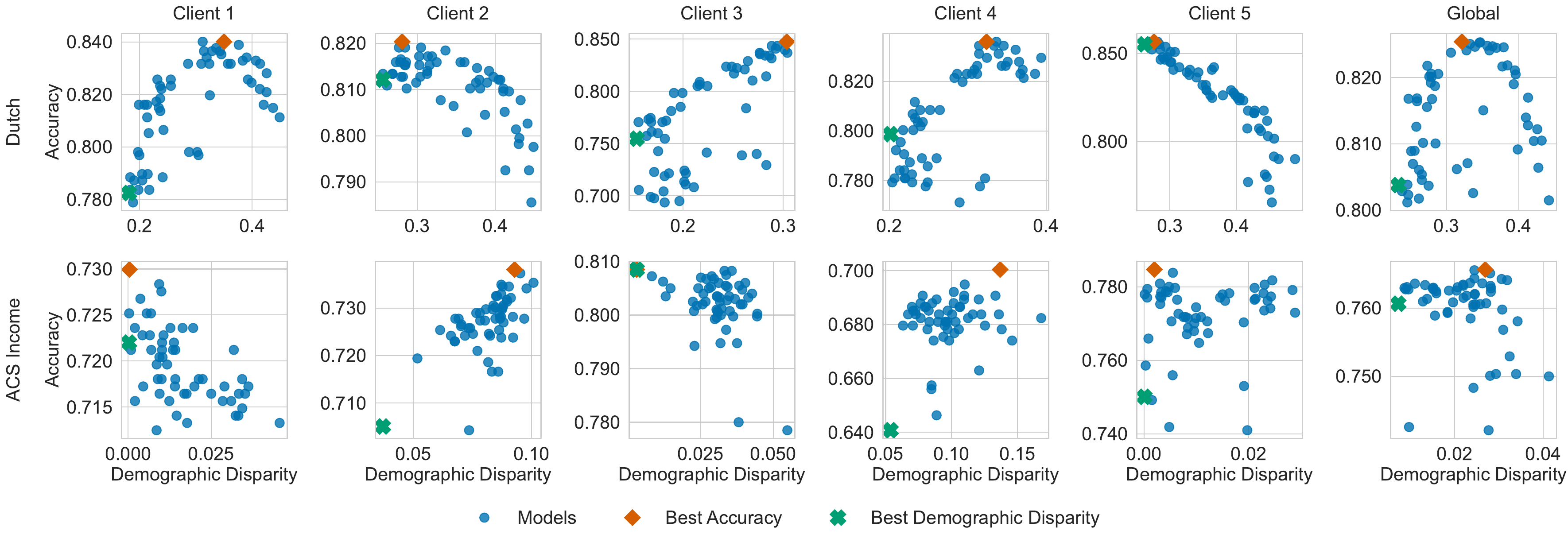}
    \caption{Demographic Disparity (on the x-axis, the lower the better) and accuracy values (on the y-axis, the higher the better) for 50 models in the $\epsilon=0.032$ global Rashomon set shown for 5 clients on the Dutch and ACS Income data.}
    \label{fig:deno_par_032}
\end{figure}

\begin{figure}
    \centering
    \includegraphics[width=\linewidth]{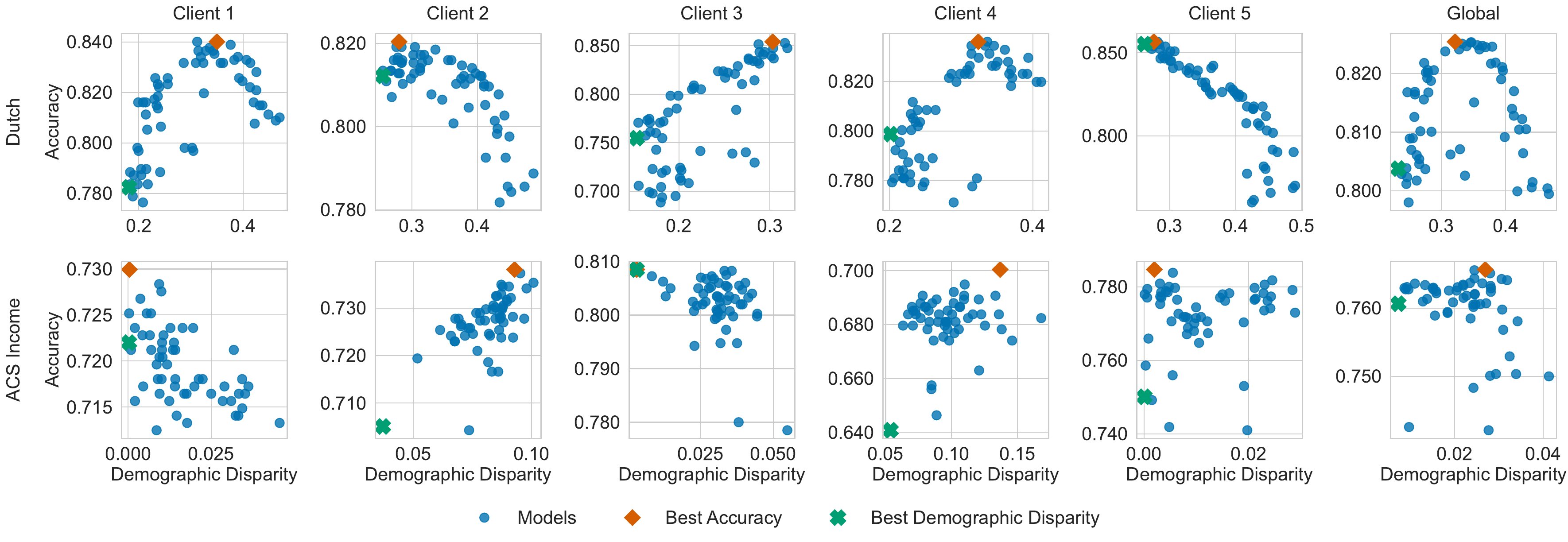}
    \caption{Demographic Disparity (on the x-axis, the lower the better) and accuracy values (on the y-axis, the higher the better) for 50 models in the $\epsilon=0.036$ global Rashomon set shown for 5 clients on the Dutch and ACS Income data.}
    \label{fig:deno_par_036}
\end{figure}

\begin{figure}
    \centering
    \includegraphics[width=\linewidth]{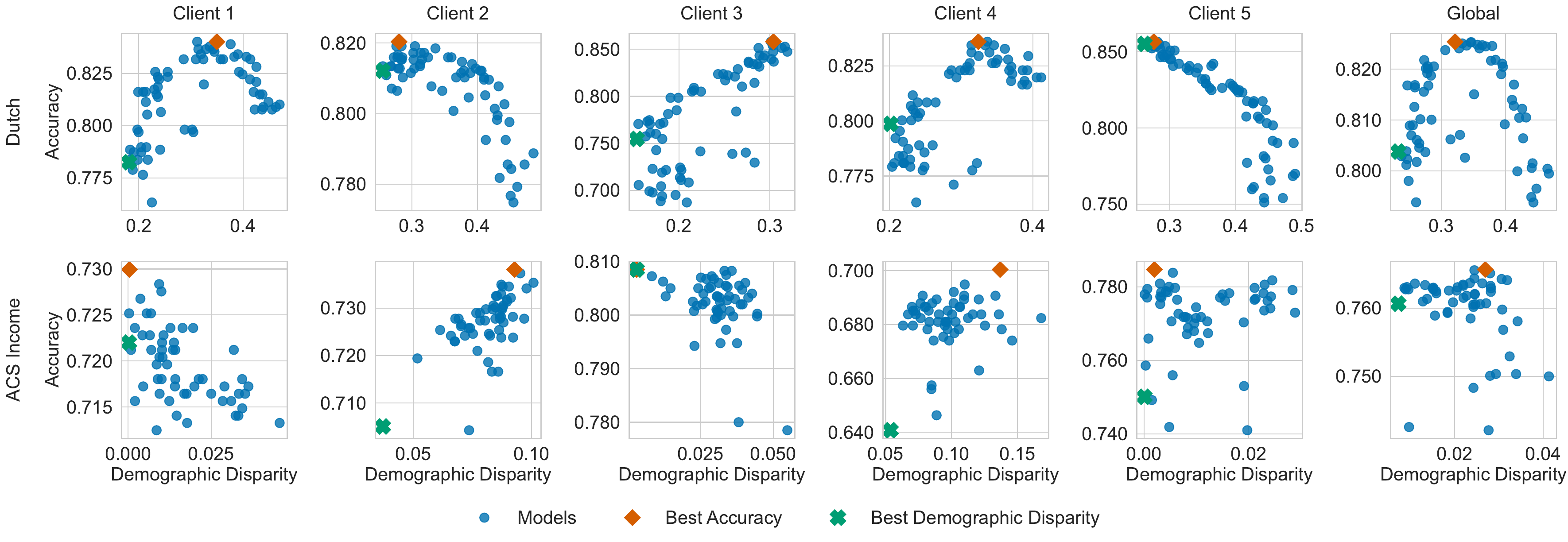}
    \caption{Demographic Disparity (on the x-axis, the lower the better) and accuracy values (on the y-axis, the higher the better) for 50 models in the $\epsilon=0.040$ global Rashomon set shown for 5 clients on the Dutch and ACS Income data.} 
    \label{fig:deno_par_040}
\end{figure}

We provide further experiments on Demographic Disparity (DD) and accuracy for 50 models sampled randomly from the global Rashomon sets for each dataset on each client's local data. 
For both Dutch and ACS Income, DD is computed with respect to \textit{sex}, which we treat as the sensitive attribute. Figure~\ref{fig:deno_par_008},~\ref{fig:deno_par_012},~\ref{fig:deno_par_016},~\ref{fig:deno_par_020},~\ref{fig:deno_par_024},~\ref{fig:deno_par_028},~\ref{fig:deno_par_032}, ~\ref{fig:deno_par_036},~\ref{fig:deno_par_040} present results for five clients on 50 models for different $\epsilon$ values. Overall, the spread of these models varies across clients, emphasizing the importance of a client-specific evaluation step in FL pipelines.


\end{document}